%% file: samplepaper.tex
\documentclass[runningheads]{llncs}

\usepackage[T1]{fontenc}
\usepackage{graphicx}
\usepackage{booktabs}
\usepackage{multirow}
\usepackage{amsmath,amssymb}
\usepackage{mathtools}
\usepackage[dvipsnames]{xcolor}
\usepackage{url}
\usepackage{hyperref}

\definecolor{DeepBrown}{RGB}{92,51,23}
\definecolor{DeepGreen}{RGB}{0,100,0}
\newcommand{\STYACC}[1]{\textcolor{DeepBrown}{#1}}
\newcommand{\STYAUC}[1]{\textcolor{DeepGreen}{#1}}

\newcommand{\tocheck}[1]{\textcolor{black}{#1}}
\definecolor{CarisColor}{RGB}{0,0,0}
\newcommand{\caris}[1]{\textcolor{CarisColor}{#1}}

\makeatletter
\let\sln@origcite\cite
\renewcommand{\cite}[1]{%
  \edef\sln@citekeys{\zap@space#1 \@empty}%
  \expandafter\sln@origcite\expandafter{\sln@citekeys}}
\makeatother
\newcommand{\citep}{\cite}
\newcommand{\citet}{\cite}

\begin{document}

\title{Probing, Fusion, and Trustworthiness: A Systematic Evaluation of Foundation Model Representations for Multimodal Cancer Analysis}
\titlerunning{Probing, Fusion, and Trustworthiness of FM Representation}

\author{Jingyu Hu\inst{1,2} \and
Giuseppe Tripodi\inst{1,3} \and
Reed Naidoo\inst{1,4} \and
Sarah F. McGough\inst{5} \and
Tapabrata Chakraborti\inst{1}}

\authorrunning{J. Hu, G. Tripodi, R. Naidoo, S. F. McGough, T. Chakraborti}

\institute{The Alan Turing Institute, London, United Kingdom \and
University of Bristol, Bristol, United Kingdom \and
University of Manchester, Manchester, United Kingdom \and
The Institute of Cancer Research, London, United Kingdom \and
Genentech, United States\\
\email{tchakraborty@turing.ac.uk}}

\maketitle

\begin{abstract}
Foundation models (FMs) have emerged as powerful representation extractors for medical data, yet their generalizability to datasets under distribution shift remains underexplored. This work systematically evaluates FM-based representations on a suite of computational pathology tasks across two
real-world commercial cohorts, \caris{IH-BC} and \caris{IH-NSCLC}, drawn from the licensed in-house (IH) \tocheck{oncology} dataset.
The analysis focuses on two modalities, whole-slide images and transcriptomic profiles, drawn from the \caris{IH} multimodal data.
We first benchmark unimodal probing performance across five FMs on eight downstream classification tasks, and find that image and omics representations carry complementary predictive signals.
Then we investigate whether multimodal fusion can yield additional gains over unimodal baselines by comparing three image-omics fusion strategies built on paired representations.
The trustworthiness of selected unimodal and multimodal pipelines is further assessed through conformal prediction. Our results show that FM representations achieve competitive performance on out-of-distribution data and that multimodal fusion helps mainly when no single modality dominates the signal.
Conformal prediction reveals that in the majority of cases where a point prediction fails, the true diagnosis remains recoverable within the prediction set, reinforcing the value of uncertainty-aware inference for clinical support.

\keywords{Foundation models \and Computational pathology \and Multimodal fusion \and Conformal prediction \and Uncertainty quantification.}
\end{abstract}

\section{Introduction}
\label{sec:intro}

Artificial intelligence has shown promising performance across a range of cancer diagnosis applications, from medical imaging~\citep{ardila2019end,mckinney2020international} and pathology~\citep{campanella2019clinical} to molecular and genomic interpretation~\citep{capper2018dna, jiao2020deep} and clinical outcome prediction~\citep{placido2023deep}.
A common workflow in these methods applies a modality-specific encoder to extract data representations, which are then consumed by either a unimodal or multimodal learning module for downstream prediction. The expressiveness of the encoded representation largely determines the quality of the downstream predictor, and prior work has extensively studied how encoder architecture and capacity match each data modality~\citep{geng2022multimodal,zhang2020multimodal}.

As medical datasets continue to grow in number and scale~\citep{deng2026project}, training a dedicated encoder from scratch for every new task becomes computationally expensive. 
To address this bottleneck, recent work starts to explore medical Foundation Models (FMs), which are pretrained on large medical corpora and used as training-free feature extractors across heterogeneous downstream tasks.
Both pathology FMs like CONCH~\citep{lu2024visual} and transcriptomic FMs like UCE~\citep{rosen2023uce} have reported promising transferability, though evaluations are mainly performed on public benchmarks that can overlap with or are close to the pretraining corpora of FMs.

In real-world scenarios, however, there are also datasets collected from industrial and commercial sources that, due to different data collection pipelines, follow different distributions from the public ones. Whether FMs can generalize to such unseen data remains an underexplored question.
This gap motivates our first question: \emph{whether FM representations can transfer to OOD datasets and yield reliable representations under probing.} 
We study this question using an in-house (\caris{IH}) real-world dataset of multi-cancer cases with paired whole-slide H\&E images and transcriptomic profiles.
We probe image representations from four image FMs and omics representations from three transcriptomic encoders and evaluate their performance on eight downstream classification tasks.

\vspace{-0.45em}
The evaluation on unimodal probing shows that image and omics representations carry complementary signals. This is consistent with the real-world setting where prognosis and treatment response are jointly determined by multiple modalities such as morphology, molecular state, and clinical context. Prior work has therefore started multimodal learning exploration that applies fusion strategies such as concatenation~\citep{chen2020pathomic} and cross-modal attention~\citep{chen2021mcat} to incorporate information across modalities.
We therefore ask \emph{whether image-omics fused representations can obtain additional predictive performance over the unimodal representations.} By pairing image and omics representations with three fusion strategies, 
we find multimodal fusion delivers stronger performance on some tasks.

The evaluations so far focus on predictive performance (e.g., accuracy). However, in high-stakes domains like medical diagnosis, high utility alone does not guarantee trustworthiness: a model may still be miscalibrated or unfair across demographic subgroups.
This leads us to further wonder \emph{how trustworthy are the unimodal and multimodal techniques studied above?} 
We apply conformal prediction~\cite{conformal_prediction} to assess predictive uncertainty and quantify the gap in predictive performance across subgroups. The selected pipelines achieve meaningful coverage guarantees and broadly similar performance across groups, while still revealing task-specific disparities for future work.

\begin{figure}[t]
  \centering
  \includegraphics[width=0.6\textwidth]{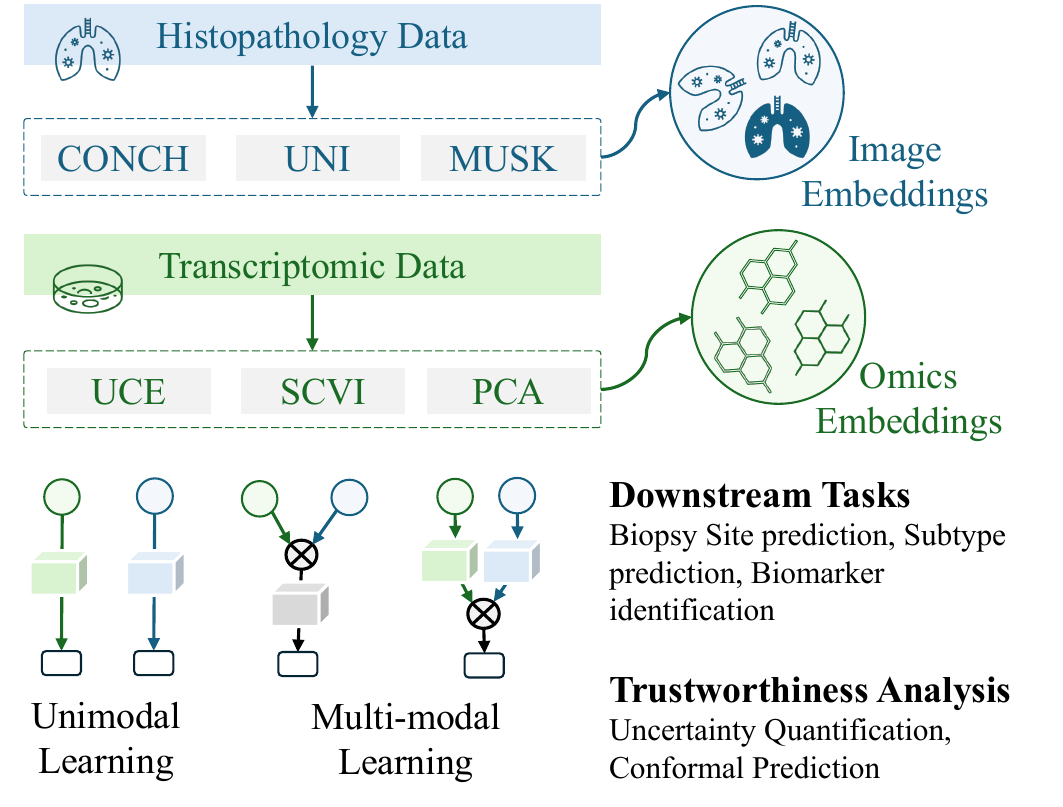}
  \caption{Overview of the Experimental Setups.}
  \label{fig:setup-overview}
\end{figure}

Figure~\ref{fig:setup-overview} illustrates the overall workflow of our work, which proceeds in three parts: unimodal probing of image and omics representations across eight downstream tasks (\S\ref{sec:rq1}), multimodal learning comparisons of three image-omics fusion strategies built on paired image-omics representations (\S\ref{sec:rq2}), and trustworthiness evaluation of selected unimodal and multimodal techniques via conformal analysis (\S\ref{sec:rq3}).

\section{Experimental Setup}
\label{sec:setup}

\textbf{Datasets Preprocessing Methods.}
\textcolor{black}{This study used the US-based deidentified Flatiron Health-Caris Life Sciences breast cancer and non-small cell lung cancer Clinical-Molecular Database (CMDB). Clinical data from the Flatiron Health Research Database \cite{flatiron_dcg, ma2020comparison, zhang2021validation} are linked to molecular data, derived from Caris Life Sciences' MI Profile\texttrademark{} comprehensive profiling in the CMDB by probabilistic matching, providing a deidentified dataset \cite{caris_miprofile, caris_micancerseek, caris_mitumorseekhybrid}.}
Based on cancer type, the collection is split into two in-house cohorts, referred to as IH-BC and IH-NSCLC. Each IH subset contains multimodal genomic data for the corresponding cancer. 
This study mainly takes two modalities of H\&E images and omics data as input, and information about cancer subtypes, biopsy sites, and biomarkers is used as the downstream prediction tasks. Specifically, \caris{IH-BC} includes LOH, Biomarker PR status, PIK3CA \tocheck{status}, Biopsy Site, and Breast Cancer Subtype identification tasks, while \caris{IH-NSCLC} includes Biopsy Site, Tumor Site and TMB identification tasks.
The datasets are split into training, calibration, validation, and test sets at a ratio of 7:3:1:1. The multimodal learning methods are trained on the training set and evaluated on the test set. More detailed preprocessing steps are reported in Appendix~\ref{sec:apx_data}.

\textbf{Modelling Methods.} 
The experiment consists of two stages: representation extraction and representation learning. For representation extraction, we evaluated four tile-level foundation models (CONCH~\citep{lu2024visual}, UNI~\citep{chen2024uni2}, Virchow~\citep{vorontsov2023virchow} and MUSK~\citep{xiang2025musk}) for H\&E WSI data and one omics foundation model (UCE~\citep{rosen2023uce}) for omics data. We also obtained omics representations using scVI~\citep{lopez2018scvi} and Principal Component Analysis (PCA)~\citep{abdi2010principal} for comparison.
In the representation learning stage, we considered five methods, each combined with different representation backbones, including two unimodal methods: H\&E image-based Multiple Instance Learning (HEMIL) and Multilayer Perceptron for omics data (GeneMLP), and three multimodal methods: concatenation-based fusion (CONTACT), Multimodal Co-Attention Transformer~\citep{chen2021mcat} (MCAT), and Late Fusion Multiple Instance Learning~\citep{reed2024latemil} (LateMIL). 
All methods are designed for the same set of tasks, and further method details are provided in Appendix~\ref{sec:apx_method}.

\textbf{Evaluation Metrics.} 
Accuracy (ACC) and AUC are used as the main evaluation metrics for model utility performance of downstream classification tasks. The ROC curve is applied to illustrate the trade-off between the true positive rate and false positive rate at different classification thresholds. More detailed formulas can be found in Appendix~\ref{sec:apx_metrics}.

\input{exp/rq1_2.tex}

\section{Experiments on Uncertainty Quantification}
\label{sec:rq3}

\begin{table*}[t]    
    \begin{minipage}{0.45\textwidth}
        \input{miscs/tables/rq3_conformal_task_latest.tex}
    \end{minipage}
    \hfill
    \begin{minipage}{0.5\textwidth}
        \input{miscs/tables/rq3_conformal_family_latest.tex}
    \end{minipage}
    
\end{table*}

Point predictions are insufficient for high-stakes oncology decision support systems, as models can still make high-confidence errors. Uncertainty quantification (UQ) addresses this by providing calibrated confidence estimates alongside each prediction~\cite{Begoli2019}. 
We analyse uncertainty using split conformal prediction (CP)~\cite{conformal_prediction}, a model-agnostic framework that wraps any trained classifier and produces a coverage-guaranteed prediction set $\mathcal{C}(x)$ rather than a single top-1 class, applied to four tasks spanning diverse label structures: LOH, Biopsy Site, Subtype, and Tumor Site. 
We report results at $\alpha = 0.10$ (90\% coverage target); additional $\alpha$ values are in Appendix~\ref{sec:apx_conf_results}.
We evaluate conformal prediction using three metrics: empirical coverage, average set size, and singleton rate. Empirical coverage measures how often the true label lies in the prediction set, while average set size and singleton rate quantify how informative and specific those sets are. Further details are provided in Appendix~\ref{sec:apx_uq}.

As shown in Table~\ref{tab:conformal_task_main}, all four multiclass tasks achieve mean coverage at or above the nominal $0.90$ target, and none exhibits a negative aggregate coverage gap. 
The two NSCLC endpoints deviate most strongly from the nominal target: NSCLC Biopsy Site and NSCLC Tumour Site have mean coverages of $0.931$ and $0.915$, respectively. 
This over-coverage is directly linked to their near-zero singleton rates; models are rarely confident enough to commit to a single class, so prediction sets routinely span two or three classes, driving coverage well above the nominal level at the cost of efficiency.
This behaviour is consistent with the intrinsic difficulty of discriminating between lung subregions in routine clinical samples, which often yields broader, more overlapping probability distributions and therefore larger, more conservative conformal sets.
The breast cancer (BC) tasks are substantially tighter. BC-LOH and BC-Subtype lie within $0.003$ of the target coverage at the aggregate level, with average set sizes of approximately $2.0$ out of three and four classes, respectively, indicating a more favourable reliability-efficiency trade-off. 
This tighter behaviour aligns with the higher predictive accuracy on BC tasks and suggests that the available image and omics features provide a stronger signal for distinguishing LOH status and molecular subtypes than for separating lung subregions, enabling models to concentrate probability mass more sharply and yield more efficient prediction sets.
The rescue rate (how often the true label is recovered when the top-1 prediction is wrong) reinforces these results. NSCLC tasks benefit most from larger sets ($87.3\%$ for Biopsy Site, $84.8\%$ for Tumour Site), while BC tasks still recover the true label in over $72\%$ of failures, confirming that conformal prediction provides a meaningful safety net across all tasks.

Table~\ref{tab:conformal_family_coverage_setsize} disaggregates these results by model family. For any given task, the results are more stable, the spread in average prediction-set size across the five families is less than $0.5$, indicating that efficiency is largely architecture-agnostic. 
On the BC tasks, where coverage is closest to the nominal target, CONTACT produces the most compact sets but falls marginally below $0.90$ on both LOH  and Subtype, with LateMIL also under-covering slightly on Subtype. 
HEMIL is the most conservative family: it attains the highest coverage on BC Subtype ($0.912$) but at the cost of the largest average set size ($2.34$), consistent with a more dispersed softmax distribution than other models.
For the NSCLC tasks, all families substantially exceed the target coverage, in line with the task-level analysis. The main source of variation is LateMIL, which reaches $0.942$ on NSCLC Biopsy Site but drops to $0.910$ on NSCLC Tumor Site, while GeneMLP is the only family to approach the lower bound on NSCLC Tumor Site. 
Taken together, these family-level patterns suggest that conformal behaviour is largely robust to the choice of backbone: architectural differences induce only modest changes in coverage and set size, which are small compared with the much larger task-specific effects.

\section{Conclusion}
We evaluated foundation model representations for multimodal cancer analysis on two real-world \caris{IH} cohorts across three parts: unimodal probing, image-omics representations fusion, and methods trustworthiness. The results show that image FMs transfer reasonably to unseen data, while the classic PCA baseline still matches or even beats omics FMs in performance. Fusion helps when both modalities contribute to the prediction, but it can underperform when the signal from one modality dominates.
CP provides an architecture-agnostic uncertainty layer in which task difficulty, rather than model choice, governs prediction-set size and coverage, with prediction sets consistently recovering the true label when the top-1 prediction fails. 
\textcolor{black}{Exploring new representation alignment methods and unified multimodal FMs to improve the performance of multimodal methods would be helpful future directions.}

\section*{Acknowledgments}
\textcolor{black}{Portions of this research were conducted with the advanced computing resources provided by the CSCoE Converge platform. We also thank the CSCoE Scientific Computing staff for technical support. We thank Cyrus Manuel and Evan Liu for generating and providing access to the image embeddings.}

\section*{Impact Statement}

By coupling predictive evaluation with conformal uncertainty quantification, the study supports a shift from single point predictions to prediction sets with marginal coverage guarantees that preserve the true diagnosis in the majority of failure cases, reducing the risk of silently discarding the correct label and supporting safer human–AI collaboration in clinician decision support.

\nocite{langley00}

\bibliographystyle{splncs04}
\bibliography{paper_refs}

\appendix
\section{Related Work}
\label{sec:apx_rl}

\subsection{Medical Foundation Models}

Foundation models (FMs) for computational pathology can be broadly categorized into tile-level and whole-slide-level approaches.
Tile-level FMs learn patch representations from histology crops, with representative examples including contrastive learning based CTransPath~\citep{wang2022CTransPath}, self-supervised model UNI~\citep{chen2024uni2}, CONtrastive learning from Captions for Histopathology (CONCH~\cite{lu2024visual}), and Virchow~\citep{vorontsov2023virchow}. 
Slide-level FMs aggregate features into holistic slide embeddings and the examples include TITAN~\citep{ding2025multimodal} and PRISM~\citep{shaikovski2024prism}.
For transcriptional FMs, UCE~\citep{rosen2023uce} learns universal cell embeddings across species, scGPT~\citep{cui2024scgpt} pretrains on over 33 million cells and Geneformer~\citep{theodoris2023geneformer,chen2026scaling} encodes rank-value gene tokens for network-level prediction.

Beyond using FMs as feature extractors, a growing line of work explores FMs as interactive generators. PathChat~\citep{lu2024pathchat} pairs a pathology vision encoder with a large language model to enable conversational diagnosis, while MedGemma~\citep{sellergren2025medgemma,sellergren2026medgemma} extends biomedical image-text reasoning to broader medical domains.
Despite these advances, recent studies~\citep{kedzierska2025zero,karasikov2025training} have also shown concerns that biomedical FMs can suffer from hallucinations, misdiagnosis, modality misalignment, and in some settings, representations that underperform simpler or classical baselines. This motivates us a closer examination of how FM representations behave on downstream classification tasks.

\subsection{Multimodal Learning for Downstream Tasks}
Multi-modal learning methods can be categorized into early fusion, intermediate fusion, late fusion, and hybrid ways by the stage of modalities combinations~\citep{krones2025mmreview,huang2020fusionsurvey,hermessi2021multimodal}. Early fusion merges raw or minimally processed inputs via feature concatenation or stacking before they enter the main model. \cite{duvieusart2022multimodal} use XGBoost on image-derived biomarkers together with ICU tabular data (vital sign values, laboratory values and metadata) for cardiomegaly classification. Another practice is to use an unsupervised encoder to compress data from different modalities (e.g., clinical records and gene expression) into a single feature vector before feeding it to a survival prediction network~\citep{cheerla2019deepearlyfuse}. Intermediate fusion first extracts modality-specific representations through separate encoders, then combines them in a shared feature space. Common examples include Pathomic Fusion~\citep{chen2020pathomic} and MCAT~\citep{chen2021mcat}. The main idea of these methods involves first learning unimodal feature representations through individual encoders, which are then fused via mechanisms to train the final model~\citep{krones2025mmreview}. Late fusion operates at the decision level, combining independent per-modality predictions through voting, averaging, or learned meta-classifiers~\citep{huang2020multimodal}. Hybrid methods mix multiple stages, with examples including PORPOISE~\citep{chen2022pan}, HEALNet~\citep{hemker2024healnet} and SurvPath~\citep{jaume2024survpath}.

\subsection{Uncertainty Quantification in Medical AI}

Deploying machine learning in high-stakes medical settings requires not only high 
predictive accuracy, but also reliable indicators of when a prediction can be 
trusted~\citep{begoli2019need}.
Early work applied approximate Bayesian methods, notably Monte Carlo dropout and deep ensembles, to clinical imaging, demonstrating that uncertainty estimates can flag unreliable predictions across a range of tasks, including diabetic retinopathy screening~\citep{leibig2017leveraging}, brain lesion segmentation~\citep{roy2019bayesian}, and lung cancer and skin lesion analysis~\citep{Zahari2023, enhancing_nodule_risk_uncertainty}.
Calibration has emerged as a complementary concern; models trained with cross-entropy are systematically overconfident, making post-hoc recalibration a standard practice~\citep{guo2017calibration}. 

Conformal prediction (CP)~\citep{conformal_prediction} provides finite-sample, distribution-free coverage guarantees without assumptions on the underlying model or data-generating process, making it well suited to heterogeneous clinical datasets. 
Recent work has extended CP from classification to risk-bounded interval estimation; \citet{angelopoulos2022riskcontrol} shows how multimodal survival models can be wrapped to yield prediction intervals with a bounded false-coverage rate, and \citet{Dey_Banerji_Basuchowdhuri_Saha_Parashar_Chakraborti_2025} applies CP to PathGen-based multimodal predictors to obtain calibrated grade sets and survival-risk intervals in computational pathology and transcriptomics.

\section{Implementation Details}
This appendix details the implementation of our benchmark of data preparation and exploratory analysis of \caris{IH} data (\S\ref{sec:apx_data}), representation extraction and learning methods used in experiments (\S\ref{sec:apx_method}), evaluation metrics for utility performance (\S\ref{sec:apx_metrics}), and conformal prediction for uncertainty quantification (\S\ref{sec:apx_uq}).

\input{apx/apx_data_method_metrics.tex}

\input{apx/apx_conformal.tex}

\section{Appendix: Full Results}

Table \ref{tab:rq2} shows detailed results for all multimodal techniques with different image and omics representations backbones.

Overall, models on BC tasks show better predictive performance than NSCLC tasks. 
This difference may reflect multiple factors such as task definition, label noise, and demographic composition. A subgroup analysis would be needed to isolate demographic effects: the BC models are effectively built on a more demographically homogeneous population than the NSCLC models.
Due to the nature of the disease, BC dataset we extracted are all female patients, so the model is trained and evaluated on a single-gender cohort. In contrast, NSCLC dataset contains a more balanced distribution of male and female patients, but demographic information is excluded from modelling.
Similar clinical presentations can correspond to different classification labels across genders, making the prediction task inherently more challenging.
Incorporating demographic information into the modelling process and evaluating variations in predictive performance across subgroups would be beneficial for building more robust predictive models.

\tocheck{Across different downstream tasks, we observe a pattern that task labels anchored in tissue context tend to favor image representations, whereas task labels anchored in molecular state tend to favor omics representations. For instance, the image-based HEMIL probe overall outperforms the omics-based GeneMLP probe on the Biopsy Site identification tasks for both NSCLC and BC datasets. This may be because biopsy site is primarily determined by surrounding tissue context and tumor morphology, both of which favor image-based probing. By contrast, biomarker-status tasks, such as BC-PR, target molecular states whose primary substrate is transcriptional, and therefore favor omics-leaning probing. As a result, GeneMLP outperforms HEMIL on BC-PR status prediction.}
\input{miscs/tables/rq2_multimodal_table.tex}

\begin{figure}[t]
  \centering
  \includegraphics[width=\textwidth]{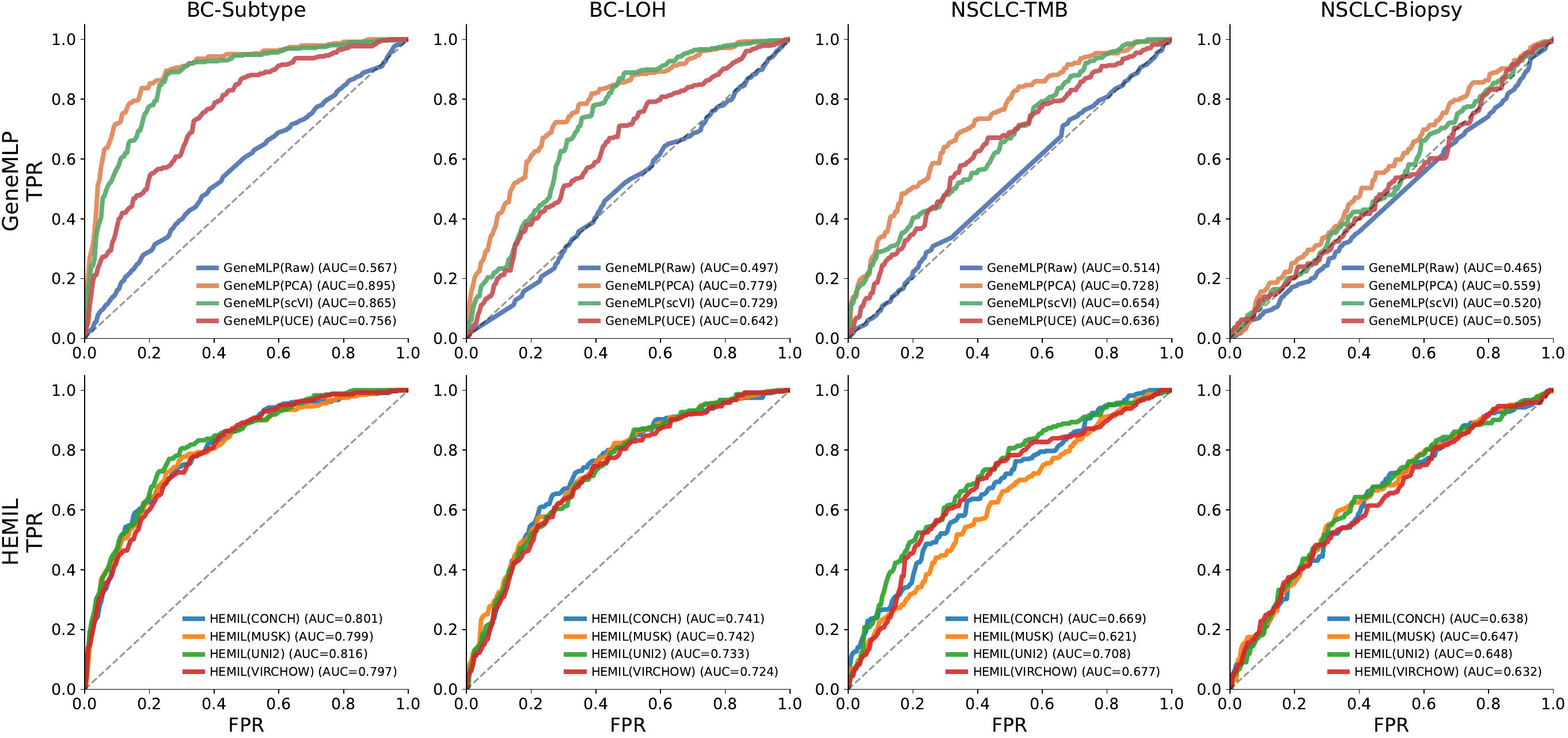}
  \caption{ROC curves of unimodal baselines across different tasks (BC-Subtype, BC-LOH, NSCLC-TMB, NSCLC-Biopsy) Here `RAW' refers to directly feeding the full set of gene TPM values as omics features. }
  \label{fig:all-roc}
\end{figure}

Figure \ref{fig:all-roc} reports macro-averaged ROC curves of the two unimodal baselines (GeneMLP for omics, HEMIL for WSI) across four representative tasks under different encoder backbones. Overall, the patterns across different tasks are consistent with what we observed in the main text. The various image-based foundation models show stable predictive performance. For omics representations, all three encoded versions outperform using raw gene expression directly. That said, the omics foundation model UCE generally underperforms PCA and scVI. This is likely because UCE is pre-trained and generates representations in a training-free manner, whereas PCA and scVI are further fitted on the training set before producing representations. Fine-tuning omics foundation models on domain-specific data in future work can help them generate more informative embeddings.

\subsection{Conformal Prediction}
\label{sec:apx_conf_results}

\begin{figure}[http]
    \centering
    \includegraphics[width=0.92\linewidth]{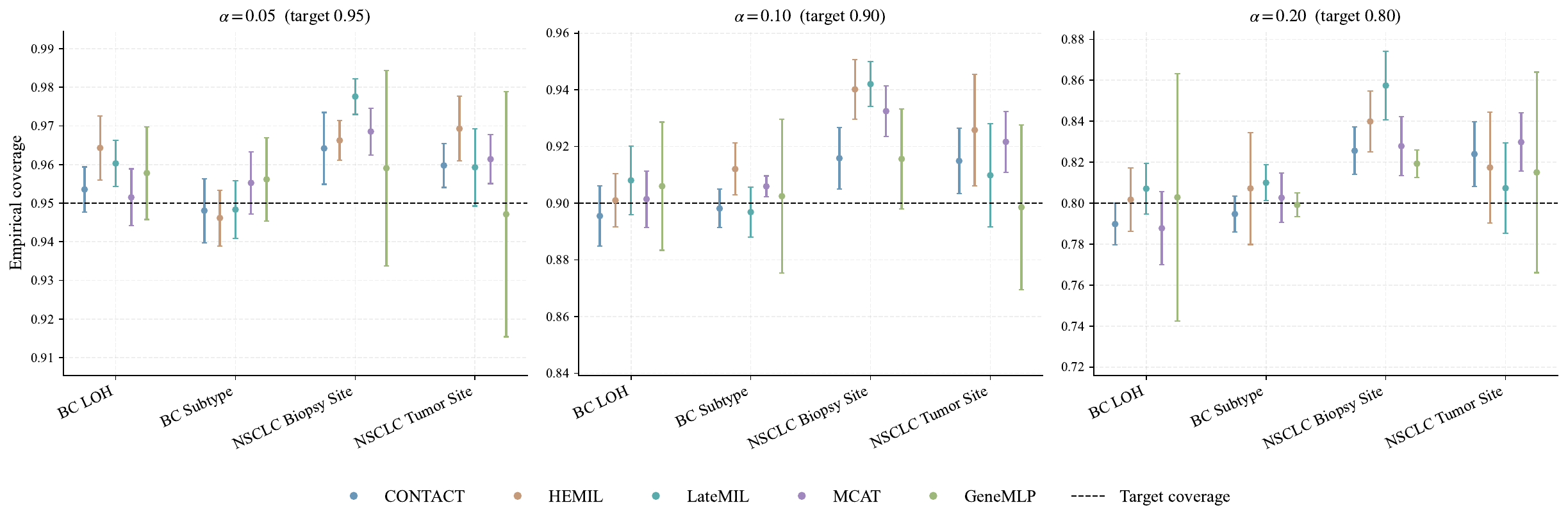}
    \caption{Empirical coverage across models and target coverage levels $\alpha \in \{0.05, 0.10, 0.20\}$. Each point is the mean coverage averaged over encoder configurations for a given model and task.}
    \label{fig:family_reliability}
\end{figure}

Figure~\ref{fig:family_reliability} extends the single-level analysis to the full range of target coverages. The same task-level structure observed at $\alpha = 0.10$ is evident across all three panels, the BC tasks track the nominal targets closely, whereas the NSCLC tasks consistently overshoot, because their harder multi‑site discrimination task leads to more variable, higher non‑conformity scores and thus larger, more conservative prediction sets.

At $\alpha = 0.05$ (target $0.95$), BC configurations cluster tightly between $0.95$ and $0.97$. The only clear under-coverage is GeneMLP on NSCLC Tumour Site (coverage $0.9471$, gap $-0.003$), while LateMIL on NSCLC Biopsy Site shows the strongest over-coverage ($0.9776$, gap $+0.028$; Table~\ref{tab:conformal_alpha_sweep}). 
At $\alpha = 0.10$ (target $0.90$), BC families again form a narrow band around the target, whereas NSCLC families overshoot by roughly 2--4 percentage points; GeneMLP on NSCLC Tumour Site ($0.8986$, gap $-0.001$) is the only near-miss. 
At $\alpha = 0.20$ (target $0.80$), NSCLC over-coverage remains high, while CONTACT and MCAT on BC LOH dip slightly below target (gaps $-0.010$ and $-0.012$), making this level the most prone to under-coverage on BC tasks (Table~\ref{tab:conformal_alpha_sweep}).
The relative ordering of model families within each task remains stable and the spread of coverages does not grow systematically with $\alpha$, indicating that the qualitative conclusions from the $\alpha = 0.10$ setting generalise across the coverage sweep.

Table~\ref{tab:conformal_alpha_sweep} also reports probabilistic calibration via the expected calibration error (ECE)~\cite{ECE} and the multi-class Brier score~\cite{Dawood2023}, both of which are independent of $\alpha$ and therefore reflect intrinsic model properties rather than the choice of coverage level. 
CONTACT and MCAT are consistently well calibrated,  maintaining ECE values in the range $0.04$--$0.07$ across all four tasks, whereas LateMIL and GeneMLP are notably less reliable, with ECE reaching $0.144$ and $0.160$ on the BC tasks and peaking at $0.204$ for GeneMLP on NSCLC Tumour Site. 
The Brier score follows the same path. NSCLC tasks have substantially higher values ($0.62$--$0.77$) than BC tasks ($0.46$--$0.59$), consistent with their increased difficulty, and within each endpoint, the ordering of families matches the ECE ranking, with CONTACT and MCAT achieving the lowest Brier scores and GeneMLP the highest.

These probabilistic calibration mirrors the conformal results. Models with broad, poorly concentrated softmax distributions, as indicated by high ECE and Brier, require larger $\hat{q}$ thresholds to include the true class, increasing prediction-set sizes and producing over-coverage. 
This is most evident for GeneMLP on NSCLC Tumour Site, where high ECE ($0.204$) and Brier ($0.773$) coincide with the largest prediction sets in that task (average size $\approx 2.58$--$2.77$ across $\alpha$ levels; Table~\ref{tab:conformal_alpha_sweep}) and with the only near under coverage at $\alpha = 0.05$ (coverage $0.947$, gap $-0.003$). 
Conversely, MCAT on NSCLC Biopsy Site combines the lowest ECE in that task ($0.044$) with the smallest prediction sets, illustrating how better probabilistic calibration translates directly into tighter and more informative prediction sets.

\input{miscs/tables/full_conformal_table}

\begin{figure}[http]
    \centering
    \includegraphics[width=0.92\linewidth]{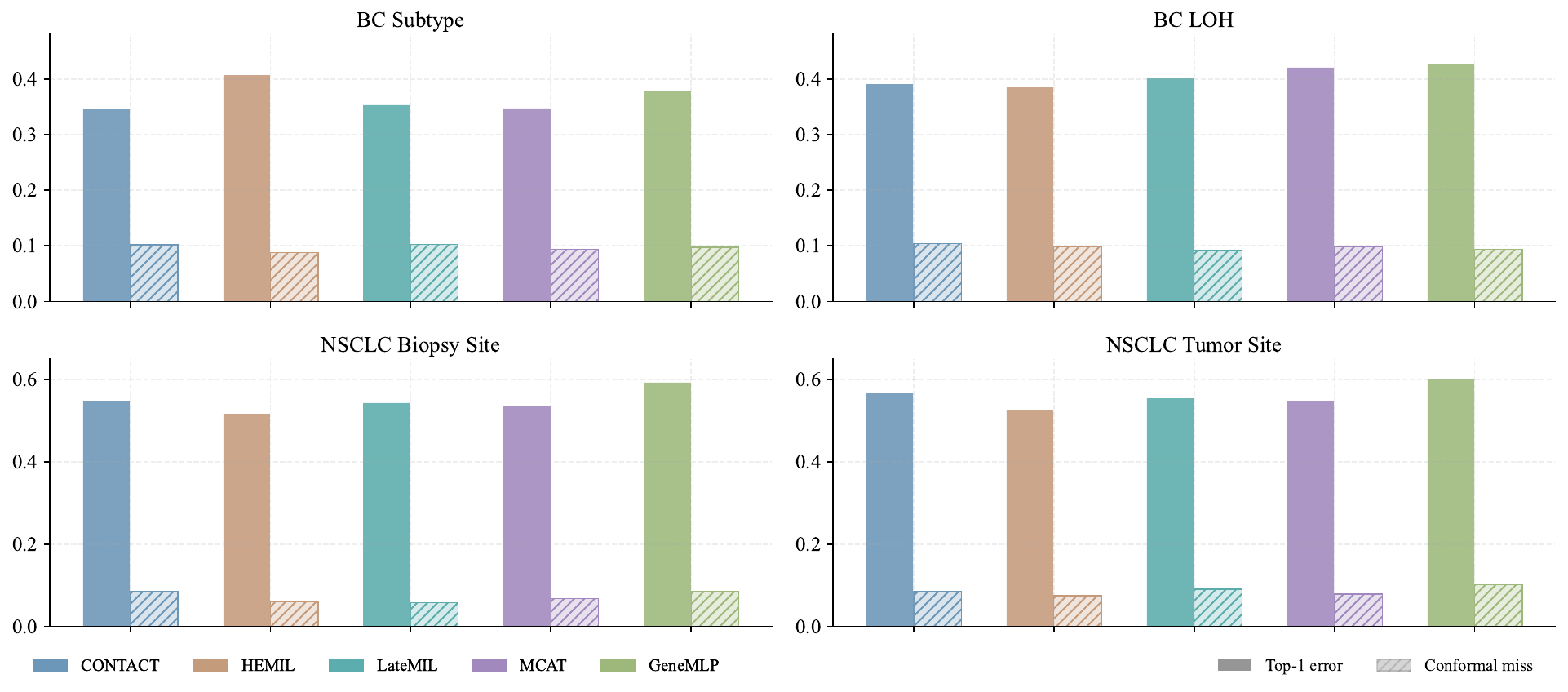}
    \caption{Error decomposition by task and model family. Each panel shows one of the four multiclass tasks; within each panel, bars are grouped by model family (colour). Solid bars represent the top-1 classification error ($1 - \text{accuracy}$); hatched bars represent the conformal miss rate ($1 - \text{coverage}$) at $\alpha=0.10$.
    The gap between the two bars quantifies the uncertainty absorbed by the prediction set, samples misclassified by the point predictor that are nonetheless covered by $\mathcal{C}(x)$ because the true class retains a softmax score above $1-\hat{q}$.}

    \label{fig:error_decomposition}
\end{figure}

Figure~\ref{fig:error_decomposition} illustrates the top-1 error, which counts how often the single most likely class is wrong, and the conformal miss rate, which counts how often the true label falls outside the prediction set $\mathcal{C}(x)$. 
The difference between these two reflects the uncertainty absorbed by the prediction set, where the classifier misclassifies the top label but the conformal wrapper still includes the true class.
Across the BC tasks (top row), top-1 error ranges from approximately $35\%$ to $43\%$ depending on the model family, yet the conformal miss rate remains close to the nominal $10\%$ target for all five families. 
The hatched bars are short and nearly identical, indicating that the choice of architecture has virtually no impact on coverage reliability at this level.
The NSCLC tasks (bottom row) show a different pattern. Top-1 error increases to $52$--$60\%$, reflecting the harder multi-site discrimination, while conformal miss rates fall below the target to about $7$--$8\%$, consistent with the systematic over-coverage reported in Table~\ref{tab:conformal_task_main}. 
The gap between solid and hatched bars is therefore largest on the NSCLC tasks, the conformal wrapper absorbs substantially more uncertainty, covering many samples that the point predictor misclassifies. 
This gap is also highly consistent across model families, suggesting that the score distributions of all architectures are broad enough that the true label typically remains inside $\mathcal{C}(x)$ regardless of backbone choice.

From the rescue rates in Table~\ref{tab:conformal_alpha_sweep}, we observe that across all four multiclass tasks and all model configurations, $79.7\%$ of misclassified samples were \emph{rescued}, meaning that the conformal set still contained the correct label even when the single-best prediction was wrong.
The rescue rate varied by clinical task: $72.6\%$ for BC~Subtype, $75.7\%$ for BC~LOH, $87.3\%$ for NSCLC Biopsy Site, and $84.8\%$ for NSCLC Tumour Site.
These numbers indicate that, in the large majority of failure cases, a hard classifier would commit to an incorrect label, whereas conformal prediction would still flag the true diagnosis as a plausible outcome and defer the final decision to a human reviewer.

Two patient-level examples from the BC~Subtype task illustrate the clinical impact of this behaviour. 
In the first example, a HEMIL model assigned a softmax probability of $0.368$ to HR+/HER2-Low and $0.364$ to the true class HR+/HER2-Negative, a very small margin of $\Delta p = 0.004$. 
A standard classifier would pick HR+/HER2-Low and be wrong.\footnote{See, for example, the FDA approval summary for fam-trastuzumab deruxtecan-nxki (Enhertu) in HER2-low breast cancer: \url{https://www.fda.gov/drugs/resources-information-approved-drugs/fda-approves-fam-trastuzumab-deruxtecan-nxki-her2-low-breast-cancer}.}
A false HR+/HER2-Low label would therefore wrongly suggest that the patient is eligible for T-DXd and could send them down an inappropriate treatment pathway. By contrast, the conformal prediction set at $\hat{q} = 0.860$ was $\mathcal{C}(x)=\{\text{HR+/HER2-Low},\,\text{HR+/HER2-Neg},\,$ $\text{Indeterminate}\}$, a three-class output that clearly shows HER2 status is uncertain and that additional IHC or FISH testing is needed before deciding on treatment.

In the second example, the same model assigned a probability of $0.336$ to Indeterminate and $0.331$ to the true class, Triple Negative, again a tiny margin ($\Delta p = 0.004$). 
Triple Negative Breast Cancer (TNBC) is defined by the absence of some receptors and is managed very differently from an Indeterminate case. Patients with TNBC may receive intensive systemic therapies such as chemotherapy. 
An Indeterminate prediction would instead trigger further biomarker workup and delay the start of TNBC-directed therapy, which is problematic given the aggressive course of the disease~\cite{Bagegni2022}.
The conformal prediction set $\mathcal{C}(x)=\{\text{HR+/HER2 Low},\,\text{Indeterminate},\,\text{Triple Neg.}\}$ keeps Triple Negative as an active diagnostic option alongside Indeterminate, encouraging clinicians to pursue both possibilities in parallel rather than committing to a single ambiguous label.
Taken together, these examples show that conformal prediction sets do more than improve statistical coverage: they translate model uncertainty into concrete clinical prompts for individual patients.

\end{document}

%% file: exp/rq1_2.tex
\section{Experiments on Unimodal Probing}
\label{sec:rq1}

\input{miscs/tables/wo_sa/rq1_unimodal_table.tex}

Table~\ref{tab:rq1} summarizes the unimodal probing performance of four types of tiles representations and three omics representations across eight downstream tasks.

Image foundation models achieve broadly comparable performance, though results vary considerably across \tocheck{cancer types}: AUC exceeds $0.9$ on \caris{IH-BC} Biopsy Site across all image FMs, whereas \caris{IH-NSCLC} Biopsy Site AUC ranges only from $0.6325$ to $0.6478$.
Among tile representations, differences between image FMs are small relative to differences across tasks, indicating that downstream task difficulty is the key factor shaping unimodal probing performance.
Omics representations provide strong unimodal signal on several tasks, and in some cases outperform image-based methods. For example, the highest accuracy on the task of BC-PIK3CA among image-based approaches is 0.7533, whereas the best omics-based result reaches 0.7933.
The classical PCA baseline generally outperforms learned transcriptomic encoders on most task and achieves highest AUC on LOH ($0.7794$), PR ($0.8159$), PIK3CA ($0.7921$), Subtype ($0.8955$), and TMB ($0.7277$).

\begin{figure}[t]
  \centering
  \includegraphics[width=0.8\linewidth]{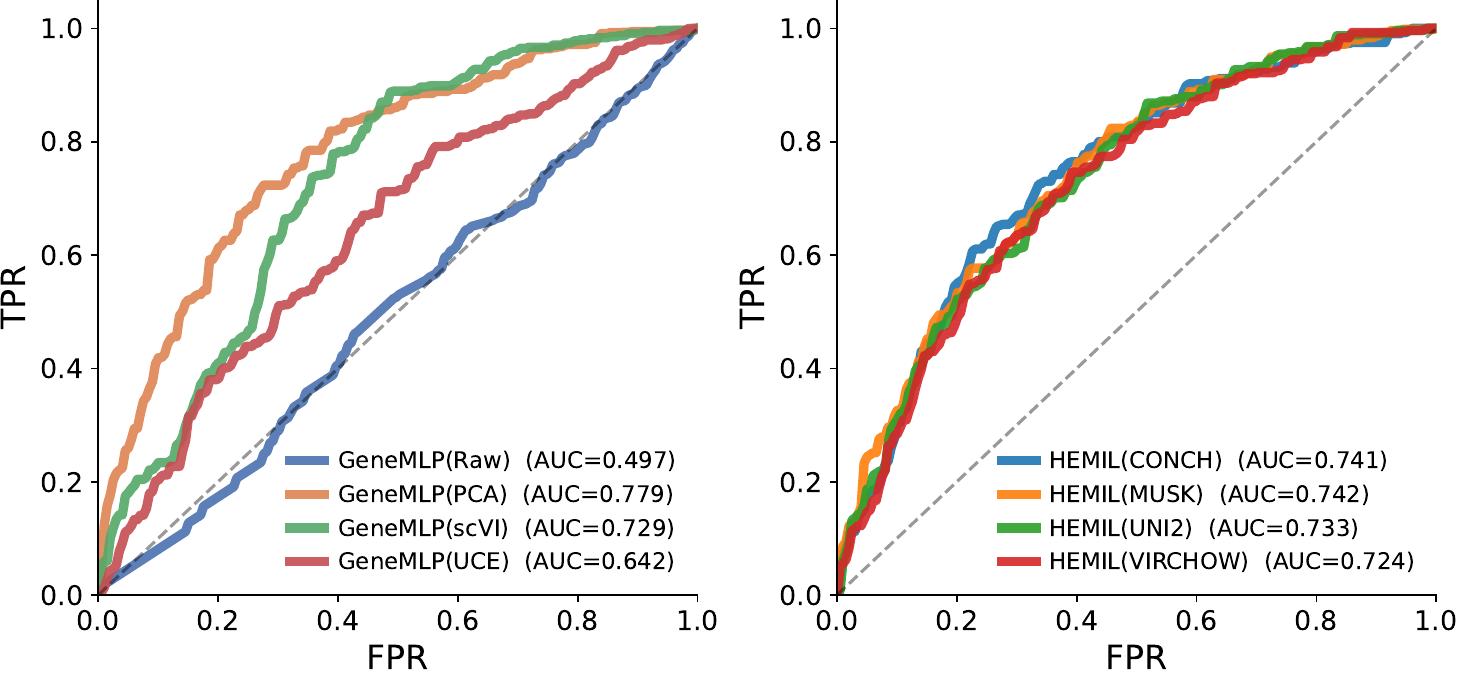}
  \caption{ROC Comparison on BC-LOH Task.}
  \label{fig:roc}
\end{figure}

We also compared ROC curves (Figure \ref{fig:roc}) and observed a consistent pattern: image representations show more stable performance with relatively small variance across FMs. All three omics representation methods (UCE, PCA, and scVI) outperform direct modeling on the raw full gene expression profile on BC-LOH. However, the foundation model UCE underperforms the non-foundation-model approaches scVI and PCA. The pattern observed in the industrial \caris{IH} dataset aligns with prior discussions on public datasets, suggesting that PCA can be better suited for capturing biological perturbations than existing omics foundation models~\citep{bendidi2024benchmarking}. This indicates that building effective transcriptomic foundation models remains an open challenge.

\section{Experiments on Multimodal Fusion }
\label{sec:rq2}

\begin{figure*}[t]
  \centering
  \includegraphics[width=\linewidth]{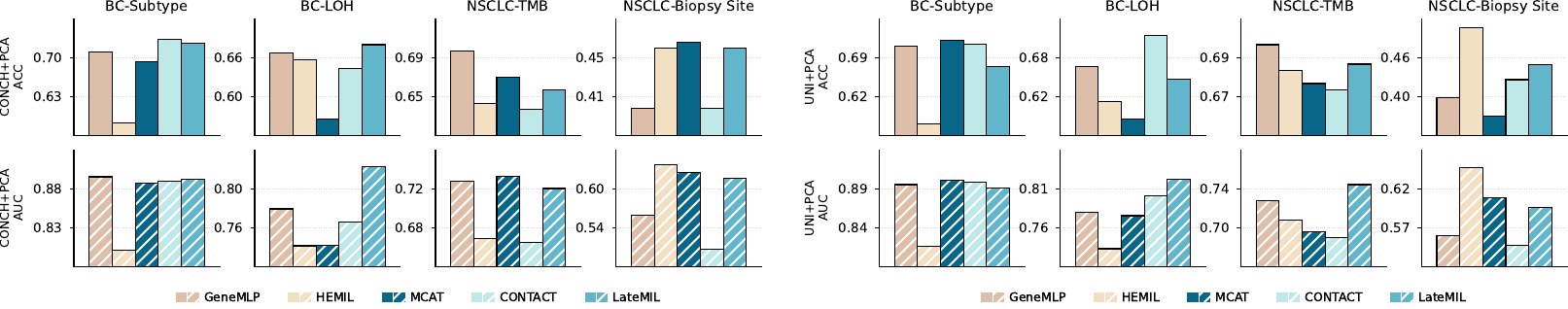}
  \caption{Performance Comparisons of Unimodal (\textcolor[HTML]{8b5e34}{GeneMLP, HEMIL}) and Multimodal (\textcolor[HTML]{124559}{MCAT, CONTACT, LateMIL}) Methods}
  \label{fig:hist}
\end{figure*}

The above results show that omics and image modalities each have strengths on different tasks, we next investigate whether fusing their representations yields additional benefit. Figure~\ref{fig:hist} compares unimodal and multimodal performance when CONCH is used as the image backbone and PCA/SCVI as the omics backbone. The complete results can be found in Table~\ref{tab:rq2} in the Appendix.

Between different fusion strategies, LateMIL is more consistent across tasks than CONTACT and MCAT. Their comparison with unimodal methods is mixed: multimodal fusion outperforms the best unimodal baseline on some tasks, but there are also cases where it shows no appreciable gain or even underperforms a unimodal model. For instance, LateMIL achieve the highest AUC on BC-LOH (under CONCH+PCA) and LateMIL remains competitive on the NSCLC-Biopsy Site task. 
On BC-Subtype task, methods reach broadly comparable accuracy and AUC, with fusion offering marginal gains over the strongest unimodal baseline.
In contrast, on the NSCLC-TMB task, GeneMLP (PCA) attains the highest ACC, with MCAT and LateMIL falling notably below it. This suggests that when a single modality carries the dominant predictive signal, fusion can weaken rather than strengthen the representation. The same pattern holds under UNI+PCA, where the relative ranking of unimodal and multimodal methods again varies across tasks. These results indicate that multimodal fusion is not universally beneficial and its utility depends on the relative informativeness of each modality for the target task. 

%% file: miscs/tables/wo_sa/rq1_unimodal_table.tex
\begin{table*}[t]
\centering
\caption{Unimodal Probing Performance on \caris{IH-BC} and \caris{IH-NSCLC} Tasks. We report both \STYACC{ACC} and \STYAUC{AUC}.}
\label{tab:rq1}
\scriptsize
\setlength{\tabcolsep}{3.5pt}
\renewcommand{\arraystretch}{1.12}
\resizebox{0.95\textwidth}{!}{%
\begin{tabular}{ll|cccc|ccc}
\hline
 & & \multicolumn{4}{c|}{HEMIL} & \multicolumn{3}{c}{GeneMLP} \\
Dataset & Task & CONCH & UNI & MUSK & Virchow & UCE & SCVI & PCA \\
\hline
\multirow{5}{*}{BC} & Biopsy Site & \STYACC{0.8780} / \STYAUC{0.9111} & \STYACC{0.8814} / \STYAUC{0.9343} & \STYACC{0.8508} / \STYAUC{0.9191} & \STYACC{0.8915} / \STYAUC{0.9412} & \STYACC{0.6136} / \STYAUC{0.7000} & \STYACC{0.7119} / \STYAUC{0.7590} & \STYACC{0.7797} / \STYAUC{0.8467} \\
 & LOH & \STYACC{0.6566} / \STYAUC{0.7414} & \STYACC{0.6128} / \STYAUC{0.7330} & \STYACC{0.6431} / \STYAUC{0.7420} & \STYACC{0.6229} / \STYAUC{0.7240} & \STYACC{0.5253} / \STYAUC{0.6419} & \STYACC{0.4882} / \STYAUC{0.7287} & \STYACC{0.6667} / \STYAUC{0.7794} \\
 & PR & \STYACC{0.6933} / \STYAUC{0.7288} & \STYACC{0.7133} / \STYAUC{0.7139} & \STYACC{0.6900} / \STYAUC{0.7453} & \STYACC{0.7267} / \STYAUC{0.7285} & \STYACC{0.6200} / \STYAUC{0.6297} & \STYACC{0.7333} / \STYAUC{0.7633} & \STYACC{0.7700} / \STYAUC{0.8159} \\
 & PIK3CA & \STYACC{0.7500} / \STYAUC{0.7355} & \STYACC{0.7433} / \STYAUC{0.7299} & \STYACC{0.7533} / \STYAUC{0.6969} & \STYACC{0.7500} / \STYAUC{0.7086} & \STYACC{0.6967} / \STYAUC{0.6097} & \STYACC{0.7033} / \STYAUC{0.7718} & \STYACC{0.7933} / \STYAUC{0.7921} \\
 & Subtype & \STYACC{0.5824} / \STYAUC{0.8013} & \STYACC{0.5714} / \STYAUC{0.8159} & \STYACC{0.5788} / \STYAUC{0.7986} & \STYACC{0.5678} / \STYAUC{0.7966} & \STYACC{0.5311} / \STYAUC{0.7558} & \STYACC{0.6520} / \STYAUC{0.8649} & \STYACC{0.7106} / \STYAUC{0.8955} \\
\hline
\multirow{3}{*}{NSCLC} & Biopsy Site & \STYACC{0.4602} / \STYAUC{0.6379} & \STYACC{0.5057} / \STYAUC{0.6478} & \STYACC{0.4830} / \STYAUC{0.6475} & \STYACC{0.4716} / \STYAUC{0.6325} & \STYACC{0.4375} / \STYAUC{0.5053} & \STYACC{0.3295} / \STYAUC{0.5205} & \STYACC{0.3977} / \STYAUC{0.5594} \\
 & TMB & \STYACC{0.6433} / \STYAUC{0.6691} & \STYACC{0.6833} / \STYAUC{0.7081} & \STYACC{0.6633} / \STYAUC{0.6210} & \STYACC{0.6500} / \STYAUC{0.6769} & \STYACC{0.6433} / \STYAUC{0.6364} & \STYACC{0.5967} / \STYAUC{0.6542} & \STYACC{0.6967} / \STYAUC{0.7277} \\
 & Tumor Site & \STYACC{0.4982} / \STYAUC{0.6283} & \STYACC{0.4912} / \STYAUC{0.6520} & \STYACC{0.4841} / \STYAUC{0.6177} & \STYACC{0.4982} / \STYAUC{0.6575} & \STYACC{0.4664} / \STYAUC{0.6100} & \STYACC{0.4629} / \STYAUC{0.6014} & \STYACC{0.4346} / \STYAUC{0.5915} \\
\hline
\end{tabular}%
}
\end{table*}

%% file: miscs/tables/rq3_conformal_task_latest.tex

\centering
\caption{Task-level conformal performance at $\alpha=0.10$, averaged over all models.
}
\label{tab:conformal_task_main}
\scriptsize
\setlength{\tabcolsep}{3pt}
\renewcommand{\arraystretch}{1.12}
\resizebox{\linewidth}{!}{%
\begin{tabular}{ll|ccccc|c}
\hline
Dataset & Task & Coverage & Cov.\ Gap & Avg.\ Set & Singleton & Accuracy & Rescue \\
\hline
\multirow{2}{*}{BC} & LOH     & 0.9021 & +0.0021 & 1.986 & 0.236 & 0.596 & 75.7\% \\
                    & Subtype & 0.9017 & +0.0017 & 1.995 & 0.310 & 0.641 & 72.6\% \\
\hline
\multirow{2}{*}{NSCLC} & Biopsy Site & 0.9309 & +0.0309 & 2.553 & 0.014 & 0.456 & 87.3\% \\
                       & Tumor Site  & 0.9154 & +0.0154 & 2.507 & 0.033 & 0.444 & 84.8\% \\
\hline
\end{tabular}%
}

%% file: miscs/tables/rq3_conformal_family_latest.tex
\centering
\caption{Conformal performance by model at $\alpha=0.10$. Each cell: Coverage\,/\,Avg.\ Set Size.}
\label{tab:conformal_family_coverage_setsize}
\scriptsize
\setlength{\tabcolsep}{2.5pt}
\renewcommand{\arraystretch}{1.12}
\resizebox{\linewidth}{!}{%
\begin{tabular}{cc|ccccc}
\hline
Dataset & Task & CONTACT & HEMIL & LateMIL & MCAT & GeneMLP \\
\hline
\multirow{2}{*}{BC} & LOH & 0.8955\,/\,1.971 & 0.9010\,/\,2.016 & 0.9081\,/\,1.967 & 0.9014\,/\,1.981 & 0.9060\,/\,2.079 \\
 & Subtype & 0.8981\,/\,1.870 & 0.9120\,/\,2.338 & 0.8969\,/\,1.994 & 0.9059\,/\,1.927 & 0.9025\,/\,2.111 \\
\hline
\multirow{2}{*}{NSCLC} & Biopsy Site & 0.9159\,/\,2.522 & 0.9402\,/\,2.549 & 0.9420\,/\,2.616 & 0.9325\,/\,2.493 & 0.9156\,/\,2.611 \\
 & Tumor Site & 0.9149\,/\,2.509 & 0.9258\,/\,2.481 & 0.9098\,/\,2.514 & 0.9216\,/\,2.487 & 0.8986\,/\,2.584 \\
\hline
\end{tabular}%
}


%% file: apx/apx_data_method_metrics.tex
\subsection{In-house (\caris{IH}) Datasets}
\label{sec:apx_data}

This subsection describes how \caris{IH-BC} and \caris{IH-NSCLC} cohorts are prepared for our experiments. We first specify notations and preprocessing steps, and then report exploratory statistics that summarize the distribution of the \caris{IH} dataset.
\textcolor{black}{The data that support the findings of this study were originated by and are the property of Flatiron Health, Inc. and Caris Life Sciences. Requests for data sharing by license or by permission for the specific purpose of replicating results in this manuscript can be submitted to PublicationsDataAccess@flatiron.com and cmdb-caris@flatiron.com.}

\subsubsection{Data Preprocessing}
\label{sec:data_selection_v2}

Both collected \caris{IH-BC} and \caris{IH-NSCLC} datasets follow the format $\mathcal{D} = \{(\mathbf{x}_{i,\text{img}},\, \mathbf{x}_{i,\text{omics}},\, y_i)\}_{i=1}^{N}$, describing multi-modal information from $N$ cases, each indexed by a unique \texttt{case\_id} $i \in \{1, \ldots, N\}$. Here $y_i \in \mathcal{Y}$ is the label for the downstream classification task, where $\mathcal{Y}$ denotes the task-specific label space (enumerated later in Table~\ref{tab:apx_ds_class}).
Each case $i$ is linked to corresponding raw image data $\mathbf{x}_{i,\text{img}}$ and raw omics data $\mathbf{x}_{i,\text{omics}}$. This study selects H\&E whole slide images as image data and TPM RNA gene expression as omics data.

The dataset is split into train, calibration, validation, and test sets with an approximate ratio of 7:3:1:1. Specifically, \caris{IH-BC} includes 3{,}747 unique case IDs, with 2{,}147 training, 1{,}000 calibration, 300 validation, and 300 test samples. \caris{IH-NSCLC} includes 3{,}887 unique case IDs, with 2{,}287 training, 1000 calibration, 300 validation, and 300 test samples. To keep the sampled subset consistent with the original cohorts, we apply a stratified split to preserve data distributions.

Table~\ref{tab:apx_ds_class} enumerated all eight downstream classification tasks considered in this study and their label spaces $\mathcal{Y}$ and class cardinalities $|\mathcal{Y}|$.
For each downstream task, we recompute the class label distribution on the sampled subset and exclude any class whose samples account for less than 8\% of the total, as such classes contain too few samples for the model to learn from effectively.

\input{miscs/tables/apx_ds_class.tex}



\label{sec:eda}
\subsection{Models Implementation}
\label{sec:apx_method}

This subsection details the models behind our benchmark. We first describe the frozen image and omics foundation models used to produce case-level representations (\S\ref{sec:apx_repgen}), then specify the unimodal and multimodal learning methods trained on top of these representations (\S\ref{sec:apx_replearn}), and finally list the computing resources used for all runs (\S\ref{sec:apx_compute}).

\subsubsection{Representation Generation Models}
\label{sec:apx_repgen}

The experiments use frozen representations before downstream learning. For each case $i$, the raw image $\mathbf{x}_{i,\text{img}}$ is encoded into an image representation $\mathbf{z}_{i,\text{img}} \in \mathbb{R}^{B_i \times d_{\text{img}}}$, a bag of $B_i$ tile embeddings, where each tile embedding is a $d_{\text{img}}$-dimensional vector and the bag size $B_i$ varies across cases. The raw omics data $\mathbf{x}_{i,\text{omics}}$ is encoded into a case-level omics representation $\mathbf{z}_{i,\text{omics}} \in \mathbb{R}^{d_{\text{omics}}}$. 

To obtain the image representation $\mathbf{z}_{i,\text{img}}$, each H\&E whole slide image is divided into tiles, and a pretrained pathology encoder then maps each retained tile into a feature vector, forming a bag of $d_{\text{img}}$-dimensional tile embeddings. When multiple slides are available for the same case, all matched slide bags are used during training, while one slide bag is used during validation and testing.


We extract representations from four different foundation models (CONCH, UNI, Virchow, MUSK).
CONCH-v1.5 is a pathology vision language foundation model trained on histology image and text pairs. Virchow is a large pathology foundation model trained at scale on cancer histopathology. UNI2 is a general purpose pathology encoder for H\&E tile representations. MUSK is a multimodal pathology foundation model for image and text representation learning.

We evaluate three omics representation generation methods (UCE, PCA, scVI).
UCE is a cell foundation model and is generates 1280-dimensional transcriptomic representations.
PCA is a dimensionality reduction method that maps the gene expression matrix to 2000 principal components.
scVI is a generative model based on a variational autoencoder. It is fitted with 3000 highly variable genes and provides a 1000-dimensional latent representation.

\subsubsection{Representation Learning Models}
\label{sec:apx_replearn}

\begin{figure*}[http]
    \centering
    \includegraphics[width=\linewidth]{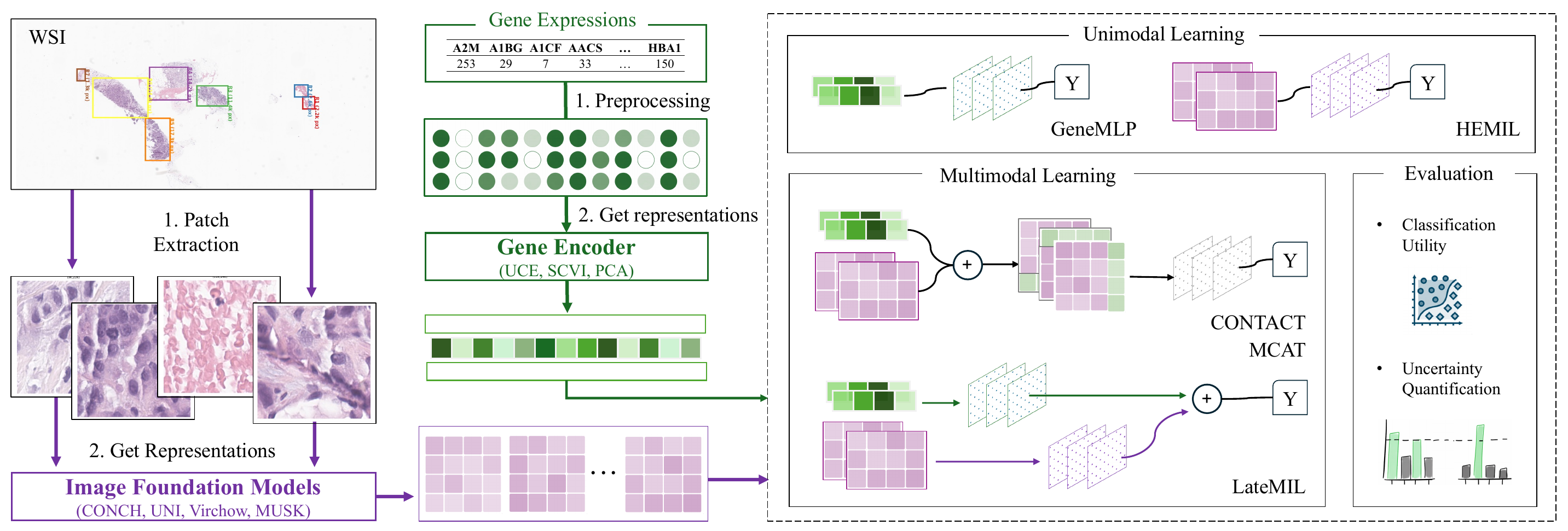}
    \caption{The Detailed Workflow}
    \label{fig:detailed_workflow}
\end{figure*}

Figure \ref{fig:detailed_workflow} presents a detailed workflow of this work. We consider five representation learning methods including two unimodal methods, HEMIL and GeneMLP, and three multimodal methods, CONTACT, MCAT, and LateMIL.
All models are trained on the training split and final results are evaluated on the test split, with the random seed fixed to 42 throughout.

\textbf{HEMIL} is an H\&E-based unimodal method that applies gated-attention multiple instance learning to the image representation bag $\mathbf{z}_{i,\text{img}}$. Each tile vector is projected by a linear layer with widths $\langle d_{\text{img}}, 256 \rangle$ followed by GELU and dropout. The attention module has two parallel linear maps of widths $\langle 256, 256 \rangle$, one with tanh and the other with sigmoid; their elementwise product is mapped to one attention logit per tile, and softmax normalizes the logits across all tiles. The bag representation is the attention-weighted sum of the projected tile vectors. The classifier head has widths $\langle 256, 128, |\mathcal{Y}| \rangle$, with LayerNorm, GELU, and dropout after the hidden layer.
HEMIL is trained with weight decay=$1\times10^{-3}$ for 100 epochs and dropout 0.4. We apply early stopping with patience=20 on validation accuracy.

\textbf{GeneMLP} is an omics-based multilayer perceptron that receives the omics representation $\mathbf{z}_{i,\text{omics}}$ and has widths $\langle d_{\text{omics}}, 512, 256, 128, |\mathcal{Y}| \rangle$, with LayerNorm, ReLU, and dropout 0.2 after each hidden linear layer.
GeneMLP is trained with weight decay=$1\times10^{-4}$ for 200 epochs. We apply early stopping with patience=20 on validation set.

\textbf{CONTACT} is a multimodal fusion strategy that concatenates omics information with every image tile embedding, then applies a fusion encoder with widths $\langle d_{\text{img}} + d_{\text{omics}},\, 512 \rangle$, with LayerNorm, GELU, and dropout after each linear layer. The fused tile bag is passed to the same MIL structure of HEMIL.
CONTACT is trained with weight decay=$1\times10^{-2}$ for 100 epochs and is applied early stopping with patience=8 on validation set.

\textbf{MCAT} is a co-attention fusion model that exchanges information between image and omics tokens before MIL pooling. The image input projection maps each tile with widths $\langle d_{\text{img}}, 512 \rangle$ using LayerNorm, GELU, and dropout; the omics input projection maps $\mathbf{z}_{i,\text{omics}}$ of width 512. Learnable modality embeddings are added to both streams, and learnable positional embeddings are added to the omics tokens. The encoder stacks 2 co-attention blocks: in each block, image tokens attend to omics tokens via multi-head attention, and omics tokens attend back via another multi-head attention (attention dropout 0.1); tile and omics self-attention are disabled by default. Each stream then passes through a FFN with hidden width 2048. The final image tokens are normalized and mapped with widths $\langle 512,\, \max(d_{\text{img}}, d_{\text{omics}}) \rangle$ with LayerNorm, GELU, and dropout. The fused tile bag is classified by the same architecture of HEMIL.
MCAT is trained with weight decay=$1\times10^{-2}$ for 100 epochs, dropout 0.5, omics feature dropout 0.2, label smoothing 0.05, gradient clipping at max norm 1.0. The early stopping is applied with patience=8 on validation set.

\textbf{LateMIL} is a late-fusion variant from~\citep{reed2024latemil} that considers both WSI logits and omics logits. The WSI branch follows a dual-stream MIL structure: the instance classifier is a linear map with widths $\langle d_{\text{img}}, |\mathcal{Y}| \rangle$ producing tile-level class logits, and the bag classifier constructs a query for each tile via a two-layer  with widths $\langle d_{\text{img}}, 128, 128 \rangle$, using ReLU after the first linear layer and tanh after the second. For each class, the tile with the highest instance logit is selected as the important class tile; attention scores between all tile queries and the important class tile queries are scaled and normalized over tiles, with an identity value path by default. The class-specific bag representations are passed through a convolution with kernel size $d_{\text{img}}$ to produce WSI bag logits. The omics branch matches the architecture of GeneMLP (with LayerNorm, ReLU, and dropout 0.2). The final prediction is a fixed weighted sum of the two branch logits, with default weights 0.5.
The optimizer uses separate parameter groups with WSI learning rate $2\times10^{-4}$ and omics learning rate $2\times10^{-3}$, weight decay=$1\times10^{-4}$, for 100 epochs.

\subsubsection{Computing Resources}
\label{sec:apx_compute}

All analyses and modelling computations are performed on the data licensor's licensed compute platform. The modality learning methods are run in parallel on an HPC platform with 8 A10G GPUs with 24 GB memory each, 196 CPU cores, and 2 TB system memory.

\subsection{Evaluation Metrics}
\label{sec:apx_metrics}


Utility is evaluated on the held-out test set for each downstream classification task. We report accuracy (ACC) and area under the ROC curve (AUC) as scalar metrics, and use ROC curves to show the threshold-dependent trade-off between true positive rate (TPR) and false positive rate (FPR). Let $N$ be the number of test cases, $y_i$ the true label, $\hat{y}_i$ the predicted label, and $p_{ic}$ the predicted probability assigned to class $c$.

\paragraph{Accuracy (ACC).}
Accuracy measures the fraction of test cases whose predicted label matches the ground-truth label, $\operatorname{ACC}= \frac{1}{N}\sum_{i=1}^{N}\mathbf{1}\{\hat{y}_i = y_i\}$.

\paragraph{ROC curves.}
ROC curves evaluate probability scores over all possible classification thresholds. For a given class $c \in \mathcal{Y}$, cases with $y_i = c$ are treated as positives and all other cases are treated as negatives. At threshold $\tau$, the true positive rate (TPR) and false positive rate (FPR) are defined as

$$\operatorname{TPR}_{c}(\tau)=\frac{\sum_{i=1}^{N}\mathbf{1}\{y_i=c,\ p_{ic}\ge \tau\}}{\sum_{i=1}^{N}\mathbf{1}\{y_i=c\}},$$
$$\operatorname{FPR}_{c}(\tau)=\frac{\sum_{i=1}^{N}\mathbf{1}\{y_i\neq c,\ p_{ic}\ge \tau\}}{\sum_{i=1}^{N}\mathbf{1}\{y_i\neq c\}}.$$

The ROC curve is obtained by plotting TPR against FPR as $\tau$ varies. Binary tasks use the positive-class ROC curve. Multi-class tasks use a macro one-vs-rest ROC curve, $\overline{\operatorname{TPR}}(u)=\frac{1}{|\mathcal{Y}|}\sum_{c\in\mathcal{Y}}\operatorname{TPR}_{c}(u)$, after each class-specific ROC curve is reparameterised as $\operatorname{TPR}_c(u)$ by inverting the monotone non-increasing relation $u = \operatorname{FPR}_c(\tau)$, eliminating $\tau$.
The diagonal line $\operatorname{TPR}=\operatorname{FPR}$ is included as the random-ranking reference.

\paragraph{Area under the ROC curve (AUC).}
AUC is the area under the ROC curve and summarizes how well the model ranks positive cases above negative cases using predicted probabilities rather than hard labels. For class $c \in \mathcal{Y}$, $\operatorname{AUC}_{c}=\int_{0}^{1}\operatorname{TPR}_{c}(u)\,du$, where $u$ denotes the false positive rate. For binary tasks, AUC is computed from the positive-class probability, following the positive class specified per task in Table~\ref{tab:apx_ds_class}. For multi-class tasks, we use a one-vs-rest macro average, $\operatorname{AUC}_{\mathrm{macro}}=\frac{1}{|\mathcal{Y}|}\sum_{c\in\mathcal{Y}}\operatorname{AUC}_{c}$, where $\mathcal{Y}$ is the task label space. This gives equal weight to each class and avoids letting larger classes dominate the reported AUC.

%% file: miscs/tables/apx_ds_class.tex
\begin{table*}[t]
\centering
\caption{Statistics of \caris{IH-BC} and \caris{IH-NSCLC} downstream classification task classes. $|\mathcal{Y}|$ denotes the cardinality of the task-specific label space $\mathcal{Y}$, i.e., the number of distinct classes for that downstream task.}
\label{tab:apx_ds_class}
\scriptsize
\renewcommand{\arraystretch}{1.15}
\setlength{\tabcolsep}{6pt}
\begin{tabular}{c c c c}
\hline
Dataset & Task & $|\mathcal{Y}|$ & \parbox[c]{10.00cm}{\centering Class Names} \\
\hline
\multirow{5}{*}{\shortstack[c]{\caris{IH}\\\caris{BC}}} & Biopsy Site & 2 & \parbox[c]{10.00cm}{\centering Metastatic / Primary} \\
 & Subtypes & 4 & \parbox[c]{10.00cm}{\centering HR-Positive HER2-Low / HR-Positive HER2-Negative / Indeterminate / Triple Negative} \\
 & Biomarker PR & 2 & \parbox[c]{10.00cm}{\centering False / True} \\
 & Biomarker PIK3CA & 2 & \parbox[c]{10.00cm}{\centering False / True} \\
 & LOH & 3 & \parbox[c]{10.00cm}{\centering Equivocal / High / Low} \\
\hline
\multirow{3}{*}{\shortstack[c]{\caris{IH}\\\caris{NSCLC}}} & Biopsy Site & 3 & \parbox[c]{10.00cm}{\centering Lower lobe, lung / Lung, NOS / Upper lobe, lung} \\
 & Tumor Site & 3 & \parbox[c]{10.00cm}{\centering Lower lobe, lung / Lung, NOS / Upper lobe, lung} \\
 & TMB & 2 & \parbox[c]{10.00cm}{\centering TMB-High / TMB-Low} \\
\hline
\end{tabular}
\end{table*}

%% file: apx/apx_conformal.tex
\subsection{Conformal Quantification}
\label{sec:apx_uq}

We evaluate our classifiers in an uncertainty-aware setting using split conformal prediction at miscoverage levels $\alpha \in \{0.05, 0.10, 0.20\}$. For each $\alpha$, prediction sets are formed by including all classes whose nonconformity score falls below a calibrated threshold, guaranteeing marginal coverage $\Pr(y \in \mathcal{C}(x)) \geq 1 - \alpha$.

The nonconformity score for a sample $x$ and class $c$ is defined as
\begin{equation}
    s(x, c) = 1 - p_\theta(c \mid x),
\end{equation}
where $p_\theta(c \mid x)$ is the softmax probability assigned to class $c$.
On the calibration set, each score is evaluated at the true label, i.e.\ $s_i = s(x, c)$.
Given calibration scores $\{s_i\}_{i=1}^{n_{\mathrm{cal}}}$, the conformal threshold $\hat{q}$ is computed as the $q$-th quantile of the calibration scores, where
\begin{equation}
    q = \frac{\left\lceil (n_{\mathrm{cal}}+1)(1-\alpha)\right\rceil}{n_{\mathrm{cal}}}
\end{equation}
is the finite-sample corrected quantile level. A test class $c$ is then included in the prediction set if and only if
\begin{equation}
    s(x, c) \leq \hat{q}.
\end{equation}

We evaluate conformal prediction results using the following metrics.

\textbf{Coverage} measures the fraction of test samples for which the true label falls inside the prediction set, and is the primary metric for verifying that the nominal guarantee is satisfied,
\begin{equation}
    \mathrm{Coverage} = \frac{1}{n}\sum_{i=1}^{n}\mathbf{1}\{y_i \in \mathcal{C}(x_i)\}.
\end{equation}

\textbf{Target coverage} is the nominal coverage level imposed by the choice of miscoverage rate $\alpha$, and it is used as the reference against which empirical coverage is compared $\mathrm{Target} = 1-\alpha.$.

\textbf{Coverage gap} quantifies the deviation of empirical coverage from the nominal target. Positive values indicate over-coverage; negative values indicate a violation of the guarantee: $\mathrm{CoverageGap} = \mathrm{Coverage} - (1-\alpha).$

\textbf{Average prediction set size} measures prediction efficiency, how many classes must the model include, on average, to achieve the required coverage. Smaller values indicate sharper, more informative predictions:
\begin{equation}
    \mathrm{AvgSetSize} = \frac{1}{n}\sum_{i=1}^{n}\lvert \mathcal{C}(x_i)\rvert.
\end{equation}

\textbf{Singleton rate} measures the fraction of samples for which the model is confident enough to return exactly one class. High singleton rates indicate decisive, well-concentrated predictions:
\begin{equation}
    \mathrm{SingletonRate} = \frac{1}{n}\sum_{i=1}^{n}\mathbf{1}\{\lvert \mathcal{C}(x_i)\rvert = 1\}.
\end{equation}


We also report the exact achievable coverage level $\delta_{\mathrm{exact}}$, which differs from the nominal target $\delta_{\mathrm{target}} = 1 - \alpha$ due to the discreteness of the calibration quantile:
\begin{equation}
    \delta_{\mathrm{exact}} =
    \frac{\left\lceil (n_{\mathrm{cal}}+1)\,\delta_{\mathrm{target}}\right\rceil}{n_{\mathrm{cal}}+1}.
\end{equation}
The slack $\delta_{\mathrm{exact}} - \delta_{\mathrm{target}}$ quantifies the unavoidable over-coverage introduced by finite-sample rounding, while the empirical gap $\mathrm{Coverage} - \delta_{\mathrm{target}}$ measures the total deviation from the nominal target. 
Together, these diagnostics allow us to distinguish genuine coverage violations from artefacts of finite calibration sets.


\textbf{Expected Calibration Error (ECE)}~\cite{ECE} quantifies how well the model’s predicted probabilities agree with empirical accuracies. 
The ECE is then defined as
\begin{equation}
    \mathrm{ECE} = \sum_{m=1}^{M} \frac{|\beta_m|}{n} \left| \mathrm{acc}(\beta_m) - \mathrm{conf}(\beta_m) \right|,
\end{equation}
where $|\beta_m|$ is the number of samples in bin $\beta_m$ and $n$ is the total number of samples. This measures the average, sample-weighted discrepancy between predicted confidence and realised accuracy across bins.

\textbf{Brier score}~\cite{Dawood2023} provides a complementary measure of probabilistic accuracy and calibration. It is defined as
\begin{equation}
    \mathrm{Brier} = \frac{1}{N} \sum_{i=1}^{N} (p_i - y_i)^2,
\end{equation}
where $N$ is the number of samples, $p_i$ is the predicted probability vector in the multi-class case, for sample $i$, and $y_i$ is the corresponding one-hot encoded ground-truth label. Lower Brier scores indicate that the predicted probabilities are closer to the observed outcomes, reflecting better-calibrated and more accurate probabilistic predictions.

To quantify the practical advantage of conformal prediction over a single point estimate, we computed the \textbf{rescue rate}: among all test samples with an incorrect top-1 prediction, this is the fraction for which the true class still appears in the conformal prediction set ($\alpha = 0.10$). 
A high rescue rate indicates that, even when the model is wrong, the correct answer is preserved among the plausible candidates, reducing the risk of silently discarding the true diagnosis in a clinical workflow.

%% file: miscs/tables/rq2_multimodal_table.tex
\begin{table*}[http]
\centering
\caption{Multimodal Fusion Performance on \caris{IH-BC} and \caris{IH-NSCLC} Tasks. Each cell reports \STYACC{ACC} / \STYAUC{AUC}.}
\label{tab:rq2}
\scriptsize
\setlength{\tabcolsep}{2.8pt}
\renewcommand{\arraystretch}{1.12}
\resizebox{\textwidth}{!}{%
\begin{tabular}{cc|ccc|ccc|ccc}
\hline
Dataset & Img-modality & CONCH & UNI & MUSK & CONCH & UNI & MUSK & CONCH & UNI & MUSK \\
 & Omics-modality & PCA & PCA & PCA & UCE & UCE & UCE & SCVI & SCVI & SCVI \\
\hline
\multirow{5}{*}{\shortstack{BC \\ LOH}} & \texttt{GeneMLP} & \STYACC{0.6667} / \STYAUC{0.7794} & \STYACC{0.6667} / \STYAUC{0.7794} & \STYACC{0.6667} / \STYAUC{0.7794} & \STYACC{0.5253} / \STYAUC{0.6419} & \STYACC{0.5253} / \STYAUC{0.6419} & \STYACC{0.5253} / \STYAUC{0.6419} & \STYACC{0.4882} / \STYAUC{0.7287} & \STYACC{0.4882} / \STYAUC{0.7287} & \STYACC{0.4882} / \STYAUC{0.7287} \\
 & \texttt{HEMIL} & \STYACC{0.6566} / \STYAUC{0.7414} & \STYACC{0.6128} / \STYAUC{0.7330} & \STYACC{0.6431} / \STYAUC{0.7420} & \STYACC{0.6566} / \STYAUC{0.7414} & \STYACC{0.6128} / \STYAUC{0.7330} & \STYACC{0.6431} / \STYAUC{0.7420} & \STYACC{0.6566} / \STYAUC{0.7414} & \STYACC{0.6128} / \STYAUC{0.7330} & \STYACC{0.6431} / \STYAUC{0.7420} \\
 & \texttt{MCAT} & \STYACC{0.5657} / \STYAUC{0.7415} & \STYACC{0.5859} / \STYAUC{0.7755} & \STYACC{0.6027} / \STYAUC{0.7579} & \STYACC{0.6465} / \STYAUC{0.7325} & \STYACC{0.6195} / \STYAUC{0.7786} & \STYACC{0.5589} / \STYAUC{0.7272} & \STYACC{0.6364} / \STYAUC{0.7746} & \STYACC{0.6027} / \STYAUC{0.7725} & \STYACC{0.6094} / \STYAUC{0.7673} \\
 & \texttt{CONTACT} & \STYACC{0.6431} / \STYAUC{0.7659} & \STYACC{0.7138} / \STYAUC{0.8016} & \STYACC{0.6431} / \STYAUC{0.7312} & \STYACC{0.5926} / \STYAUC{0.7287} & \STYACC{0.6128} / \STYAUC{0.7203} & \STYACC{0.5758} / \STYAUC{0.7077} & \STYACC{0.5825} / \STYAUC{0.7472} & \STYACC{0.6263} / \STYAUC{0.7649} & \STYACC{0.6263} / \STYAUC{0.7568} \\
 & \texttt{LateMIL} & \STYACC{0.6801} / \STYAUC{0.8233} & \STYACC{0.6465} / \STYAUC{0.8223} & \STYACC{0.6835} / \STYAUC{0.7907} & \STYACC{0.6094} / \STYAUC{0.7356} & \STYACC{0.6835} / \STYAUC{0.7699} & \STYACC{0.6263} / \STYAUC{0.7185} & \STYACC{0.6532} / \STYAUC{0.7773} & \STYACC{0.6633} / \STYAUC{0.7835} & \STYACC{0.6330} / \STYAUC{0.7614} \\
\hline
\multirow{5}{*}{\shortstack{BC \\ Biopsy Site}} & \texttt{GeneMLP} & \STYACC{0.7797} / \STYAUC{0.8467} & \STYACC{0.7797} / \STYAUC{0.8467} & \STYACC{0.7797} / \STYAUC{0.8467} & \STYACC{0.6136} / \STYAUC{0.7000} & \STYACC{0.6136} / \STYAUC{0.7000} & \STYACC{0.6136} / \STYAUC{0.7000} & \STYACC{0.7119} / \STYAUC{0.7590} & \STYACC{0.7119} / \STYAUC{0.7590} & \STYACC{0.7119} / \STYAUC{0.7590} \\
 & \texttt{HEMIL} & \STYACC{0.8780} / \STYAUC{0.9111} & \STYACC{0.8814} / \STYAUC{0.9343} & \STYACC{0.8508} / \STYAUC{0.9191} & \STYACC{0.8780} / \STYAUC{0.9111} & \STYACC{0.8814} / \STYAUC{0.9343} & \STYACC{0.8508} / \STYAUC{0.9191} & \STYACC{0.8780} / \STYAUC{0.9111} & \STYACC{0.8814} / \STYAUC{0.9343} & \STYACC{0.8508} / \STYAUC{0.9191} \\
 & \texttt{MCAT} & \STYACC{0.8169} / \STYAUC{0.9015} & \STYACC{0.8508} / \STYAUC{0.9030} & \STYACC{0.8034} / \STYAUC{0.8754} & \STYACC{0.8610} / \STYAUC{0.9242} & \STYACC{0.8814} / \STYAUC{0.9252} & \STYACC{0.8407} / \STYAUC{0.9114} & \STYACC{0.8508} / \STYAUC{0.9205} & \STYACC{0.8712} / \STYAUC{0.9119} & \STYACC{0.8237} / \STYAUC{0.8995} \\
 & \texttt{CONTACT} & \STYACC{0.7525} / \STYAUC{0.8463} & \STYACC{0.7661} / \STYAUC{0.8412} & \STYACC{0.7424} / \STYAUC{0.8224} & \STYACC{0.8407} / \STYAUC{0.9194} & \STYACC{0.8610} / \STYAUC{0.9245} & \STYACC{0.8542} / \STYAUC{0.9039} & \STYACC{0.7898} / \STYAUC{0.8902} & \STYACC{0.8441} / \STYAUC{0.9181} & \STYACC{0.8373} / \STYAUC{0.9054} \\
 & \texttt{LateMIL} & \STYACC{0.8373} / \STYAUC{0.9028} & \STYACC{0.8102} / \STYAUC{0.9089} & \STYACC{0.8441} / \STYAUC{0.9037} & \STYACC{0.8542} / \STYAUC{0.9195} & \STYACC{0.8983} / \STYAUC{0.9424} & \STYACC{0.8373} / \STYAUC{0.9249} & \STYACC{0.8407} / \STYAUC{0.9218} & \STYACC{0.8983} / \STYAUC{0.9432} & \STYACC{0.8407} / \STYAUC{0.9321} \\
\hline
\multirow{5}{*}{\shortstack{BC \\ PR}} & \texttt{GeneMLP} & \STYACC{0.7700} / \STYAUC{0.8159} & \STYACC{0.7700} / \STYAUC{0.8159} & \STYACC{0.7700} / \STYAUC{0.8159} & \STYACC{0.6200} / \STYAUC{0.6297} & \STYACC{0.6200} / \STYAUC{0.6297} & \STYACC{0.6200} / \STYAUC{0.6297} & \STYACC{0.7333} / \STYAUC{0.7633} & \STYACC{0.7333} / \STYAUC{0.7633} & \STYACC{0.7333} / \STYAUC{0.7633} \\
 & \texttt{HEMIL} & \STYACC{0.6933} / \STYAUC{0.7288} & \STYACC{0.7133} / \STYAUC{0.7139} & \STYACC{0.6900} / \STYAUC{0.7453} & \STYACC{0.6933} / \STYAUC{0.7288} & \STYACC{0.7133} / \STYAUC{0.7139} & \STYACC{0.6900} / \STYAUC{0.7453} & \STYACC{0.6933} / \STYAUC{0.7288} & \STYACC{0.7133} / \STYAUC{0.7139} & \STYACC{0.6900} / \STYAUC{0.7453} \\
 & \texttt{MCAT} & \STYACC{0.7000} / \STYAUC{0.7341} & \STYACC{0.6700} / \STYAUC{0.7364} & \STYACC{0.6667} / \STYAUC{0.7266} & \STYACC{0.7267} / \STYAUC{0.7292} & \STYACC{0.6967} / \STYAUC{0.7106} & \STYACC{0.7100} / \STYAUC{0.7286} & \STYACC{0.7133} / \STYAUC{0.7513} & \STYACC{0.7067} / \STYAUC{0.7288} & \STYACC{0.7467} / \STYAUC{0.7763} \\
 & \texttt{CONTACT} & \STYACC{0.7167} / \STYAUC{0.7627} & \STYACC{0.6767} / \STYAUC{0.7331} & \STYACC{0.7100} / \STYAUC{0.7472} & \STYACC{0.7233} / \STYAUC{0.7154} & \STYACC{0.7000} / \STYAUC{0.7198} & \STYACC{0.7333} / \STYAUC{0.7165} & \STYACC{0.6933} / \STYAUC{0.7444} & \STYACC{0.7133} / \STYAUC{0.7527} & \STYACC{0.7133} / \STYAUC{0.7548} \\
 & \texttt{LateMIL} & \STYACC{0.7367} / \STYAUC{0.7890} & \STYACC{0.7100} / \STYAUC{0.7744} & \STYACC{0.7200} / \STYAUC{0.7666} & \STYACC{0.7200} / \STYAUC{0.7252} & \STYACC{0.6933} / \STYAUC{0.7005} & \STYACC{0.6967} / \STYAUC{0.7258} & \STYACC{0.7167} / \STYAUC{0.7611} & \STYACC{0.7300} / \STYAUC{0.7483} & \STYACC{0.7200} / \STYAUC{0.7633} \\
\hline
\multirow{5}{*}{\shortstack{BC \\ PIK3CA}} & \texttt{GeneMLP} & \STYACC{0.7933} / \STYAUC{0.7921} & \STYACC{0.7933} / \STYAUC{0.7921} & \STYACC{0.7933} / \STYAUC{0.7921} & \STYACC{0.6967} / \STYAUC{0.6097} & \STYACC{0.6967} / \STYAUC{0.6097} & \STYACC{0.6967} / \STYAUC{0.6097} & \STYACC{0.7033} / \STYAUC{0.7718} & \STYACC{0.7033} / \STYAUC{0.7718} & \STYACC{0.7033} / \STYAUC{0.7718} \\
 & \texttt{HEMIL} & \STYACC{0.7500} / \STYAUC{0.7355} & \STYACC{0.7433} / \STYAUC{0.7299} & \STYACC{0.7533} / \STYAUC{0.6969} & \STYACC{0.7500} / \STYAUC{0.7355} & \STYACC{0.7433} / \STYAUC{0.7299} & \STYACC{0.7533} / \STYAUC{0.6969} & \STYACC{0.7500} / \STYAUC{0.7355} & \STYACC{0.7433} / \STYAUC{0.7299} & \STYACC{0.7533} / \STYAUC{0.6969} \\
 & \texttt{MCAT} & \STYACC{0.7100} / \STYAUC{0.7600} & \STYACC{0.7600} / \STYAUC{0.7598} & \STYACC{0.6700} / \STYAUC{0.7478} & \STYACC{0.7367} / \STYAUC{0.7291} & \STYACC{0.7000} / \STYAUC{0.7191} & \STYACC{0.7533} / \STYAUC{0.7168} & \STYACC{0.7400} / \STYAUC{0.7531} & \STYACC{0.7033} / \STYAUC{0.7343} & \STYACC{0.7100} / \STYAUC{0.7866} \\
 & \texttt{CONTACT} & \STYACC{0.7300} / \STYAUC{0.7179} & \STYACC{0.7333} / \STYAUC{0.7615} & \STYACC{0.7533} / \STYAUC{0.7328} & \STYACC{0.7467} / \STYAUC{0.7053} & \STYACC{0.7333} / \STYAUC{0.6577} & \STYACC{0.7500} / \STYAUC{0.6834} & \STYACC{0.7467} / \STYAUC{0.7592} & \STYACC{0.7167} / \STYAUC{0.7724} & \STYACC{0.7300} / \STYAUC{0.7524} \\
 & \texttt{LateMIL} & \STYACC{0.8100} / \STYAUC{0.8161} & \STYACC{0.7733} / \STYAUC{0.8047} & \STYACC{0.7533} / \STYAUC{0.7978} & \STYACC{0.7000} / \STYAUC{0.7166} & \STYACC{0.7433} / \STYAUC{0.7175} & \STYACC{0.7433} / \STYAUC{0.6928} & \STYACC{0.7400} / \STYAUC{0.7815} & \STYACC{0.7400} / \STYAUC{0.7737} & \STYACC{0.7367} / \STYAUC{0.7791} \\
\hline
\multirow{5}{*}{\shortstack{BC \\ Subtype}} & \texttt{GeneMLP} & \STYACC{0.7106} / \STYAUC{0.8955} & \STYACC{0.7106} / \STYAUC{0.8955} & \STYACC{0.7106} / \STYAUC{0.8955} & \STYACC{0.5311} / \STYAUC{0.7558} & \STYACC{0.5311} / \STYAUC{0.7558} & \STYACC{0.5311} / \STYAUC{0.7558} & \STYACC{0.6520} / \STYAUC{0.8649} & \STYACC{0.6520} / \STYAUC{0.8649} & \STYACC{0.6520} / \STYAUC{0.8649} \\
 & \texttt{HEMIL} & \STYACC{0.5824} / \STYAUC{0.8013} & \STYACC{0.5714} / \STYAUC{0.8159} & \STYACC{0.5788} / \STYAUC{0.7986} & \STYACC{0.5824} / \STYAUC{0.8013} & \STYACC{0.5714} / \STYAUC{0.8159} & \STYACC{0.5788} / \STYAUC{0.7986} & \STYACC{0.5824} / \STYAUC{0.8013} & \STYACC{0.5714} / \STYAUC{0.8159} & \STYACC{0.5788} / \STYAUC{0.7986} \\
 & \texttt{MCAT} & \STYACC{0.6923} / \STYAUC{0.8868} & \STYACC{0.7216} / \STYAUC{0.9010} & \STYACC{0.6740} / \STYAUC{0.8867} & \STYACC{0.5751} / \STYAUC{0.8159} & \STYACC{0.5824} / \STYAUC{0.8243} & \STYACC{0.5897} / \STYAUC{0.8096} & \STYACC{0.6410} / \STYAUC{0.8755} & \STYACC{0.6557} / \STYAUC{0.8555} & \STYACC{0.6557} / \STYAUC{0.8755} \\
 & \texttt{CONTACT} & \STYACC{0.7326} / \STYAUC{0.8902} & \STYACC{0.7143} / \STYAUC{0.8988} & \STYACC{0.6813} / \STYAUC{0.8821} & \STYACC{0.5641} / \STYAUC{0.8151} & \STYACC{0.5641} / \STYAUC{0.8062} & \STYACC{0.5641} / \STYAUC{0.8040} & \STYACC{0.6630} / \STYAUC{0.8777} & \STYACC{0.6630} / \STYAUC{0.8808} & \STYACC{0.6777} / \STYAUC{0.8823} \\
 & \texttt{LateMIL} & \STYACC{0.7253} / \STYAUC{0.8923} & \STYACC{0.6740} / \STYAUC{0.8912} & \STYACC{0.6740} / \STYAUC{0.8786} & \STYACC{0.5678} / \STYAUC{0.8015} & \STYACC{0.6337} / \STYAUC{0.8418} & \STYACC{0.5311} / \STYAUC{0.7680} & \STYACC{0.6337} / \STYAUC{0.8597} & \STYACC{0.6813} / \STYAUC{0.8759} & \STYACC{0.6300} / \STYAUC{0.8458} \\
\hline
\multirow{5}{*}{\shortstack{NSCLC \\ Biopsy Site}} & \texttt{GeneMLP} & \STYACC{0.3977} / \STYAUC{0.5594} & \STYACC{0.3977} / \STYAUC{0.5594} & \STYACC{0.3977} / \STYAUC{0.5594} & \STYACC{0.4375} / \STYAUC{0.5053} & \STYACC{0.4375} / \STYAUC{0.5053} & \STYACC{0.4375} / \STYAUC{0.5053} & \STYACC{0.3295} / \STYAUC{0.5205} & \STYACC{0.3295} / \STYAUC{0.5205} & \STYACC{0.3295} / \STYAUC{0.5205} \\
 & \texttt{HEMIL} & \STYACC{0.4602} / \STYAUC{0.6379} & \STYACC{0.5057} / \STYAUC{0.6478} & \STYACC{0.4830} / \STYAUC{0.6475} & \STYACC{0.4602} / \STYAUC{0.6379} & \STYACC{0.5057} / \STYAUC{0.6478} & \STYACC{0.4830} / \STYAUC{0.6475} & \STYACC{0.4602} / \STYAUC{0.6379} & \STYACC{0.5057} / \STYAUC{0.6478} & \STYACC{0.4830} / \STYAUC{0.6475} \\
 & \texttt{MCAT} & \STYACC{0.4659} / \STYAUC{0.6250} & \STYACC{0.3693} / \STYAUC{0.6086} & \STYACC{0.4318} / \STYAUC{0.6136} & \STYACC{0.4773} / \STYAUC{0.6334} & \STYACC{0.4545} / \STYAUC{0.6378} & \STYACC{0.4716} / \STYAUC{0.6362} & \STYACC{0.4830} / \STYAUC{0.6007} & \STYACC{0.4318} / \STYAUC{0.5915} & \STYACC{0.4602} / \STYAUC{0.6146} \\
 & \texttt{CONTACT} & \STYACC{0.3977} / \STYAUC{0.5071} & \STYACC{0.4261} / \STYAUC{0.5469} & \STYACC{0.3693} / \STYAUC{0.5116} & \STYACC{0.5114} / \STYAUC{0.6537} & \STYACC{0.4773} / \STYAUC{0.6305} & \STYACC{0.4716} / \STYAUC{0.6039} & \STYACC{0.4602} / \STYAUC{0.6320} & \STYACC{0.4261} / \STYAUC{0.5886} & \STYACC{0.3920} / \STYAUC{0.6123} \\
 & \texttt{LateMIL} & \STYACC{0.4602} / \STYAUC{0.6163} & \STYACC{0.4489} / \STYAUC{0.5961} & \STYACC{0.4318} / \STYAUC{0.5494} & \STYACC{0.4602} / \STYAUC{0.6373} & \STYACC{0.4716} / \STYAUC{0.6208} & \STYACC{0.4432} / \STYAUC{0.5825} & \STYACC{0.4773} / \STYAUC{0.6151} & \STYACC{0.4830} / \STYAUC{0.6410} & \STYACC{0.3580} / \STYAUC{0.5730} \\
\hline
\multirow{5}{*}{\shortstack{NSCLC \\ TMB}} & \texttt{GeneMLP} & \STYACC{0.6967} / \STYAUC{0.7277} & \STYACC{0.6967} / \STYAUC{0.7277} & \STYACC{0.6967} / \STYAUC{0.7277} & \STYACC{0.6433} / \STYAUC{0.6364} & \STYACC{0.6433} / \STYAUC{0.6364} & \STYACC{0.6433} / \STYAUC{0.6364} & \STYACC{0.5967} / \STYAUC{0.6542} & \STYACC{0.5967} / \STYAUC{0.6542} & \STYACC{0.5967} / \STYAUC{0.6542} \\
 & \texttt{HEMIL} & \STYACC{0.6433} / \STYAUC{0.6691} & \STYACC{0.6833} / \STYAUC{0.7081} & \STYACC{0.6633} / \STYAUC{0.6210} & \STYACC{0.6433} / \STYAUC{0.6691} & \STYACC{0.6833} / \STYAUC{0.7081} & \STYACC{0.6633} / \STYAUC{0.6210} & \STYACC{0.6433} / \STYAUC{0.6691} & \STYACC{0.6833} / \STYAUC{0.7081} & \STYACC{0.6633} / \STYAUC{0.6210} \\
 & \texttt{MCAT} & \STYACC{0.6700} / \STYAUC{0.7330} & \STYACC{0.6767} / \STYAUC{0.6961} & \STYACC{0.6567} / \STYAUC{0.6723} & \STYACC{0.6667} / \STYAUC{0.6775} & \STYACC{0.6200} / \STYAUC{0.6711} & \STYACC{0.6767} / \STYAUC{0.6588} & \STYACC{0.6433} / \STYAUC{0.6604} & \STYACC{0.6267} / \STYAUC{0.6709} & \STYACC{0.6500} / \STYAUC{0.6802} \\
 & \texttt{CONTACT} & \STYACC{0.6367} / \STYAUC{0.6651} & \STYACC{0.6733} / \STYAUC{0.6897} & \STYACC{0.6733} / \STYAUC{0.6666} & \STYACC{0.6600} / \STYAUC{0.6703} & \STYACC{0.6667} / \STYAUC{0.6593} & \STYACC{0.6600} / \STYAUC{0.5893} & \STYACC{0.6567} / \STYAUC{0.6884} & \STYACC{0.6633} / \STYAUC{0.6711} & \STYACC{0.6700} / \STYAUC{0.6874} \\
 & \texttt{LateMIL} & \STYACC{0.6567} / \STYAUC{0.7204} & \STYACC{0.6867} / \STYAUC{0.7444} & \STYACC{0.6567} / \STYAUC{0.6991} & \STYACC{0.5633} / \STYAUC{0.6733} & \STYACC{0.6933} / \STYAUC{0.7190} & \STYACC{0.6567} / \STYAUC{0.6780} & \STYACC{0.6233} / \STYAUC{0.6559} & \STYACC{0.6800} / \STYAUC{0.6933} & \STYACC{0.6400} / \STYAUC{0.6568} \\
\hline
\multirow{5}{*}{\shortstack{NSCLC \\ Tumor Site}} & \texttt{GeneMLP} & \STYACC{0.4346} / \STYAUC{0.5915} & \STYACC{0.4346} / \STYAUC{0.5915} & \STYACC{0.4346} / \STYAUC{0.5915} & \STYACC{0.4664} / \STYAUC{0.6100} & \STYACC{0.4664} / \STYAUC{0.6100} & \STYACC{0.4664} / \STYAUC{0.6100} & \STYACC{0.4629} / \STYAUC{0.6014} & \STYACC{0.4629} / \STYAUC{0.6014} & \STYACC{0.4629} / \STYAUC{0.6014} \\
 & \texttt{HEMIL} & \STYACC{0.4982} / \STYAUC{0.6283} & \STYACC{0.4912} / \STYAUC{0.6520} & \STYACC{0.4841} / \STYAUC{0.6177} & \STYACC{0.4982} / \STYAUC{0.6283} & \STYACC{0.4912} / \STYAUC{0.6520} & \STYACC{0.4841} / \STYAUC{0.6177} & \STYACC{0.4982} / \STYAUC{0.6283} & \STYACC{0.4912} / \STYAUC{0.6520} & \STYACC{0.4841} / \STYAUC{0.6177} \\
 & \texttt{MCAT} & \STYACC{0.4841} / \STYAUC{0.6274} & \STYACC{0.4770} / \STYAUC{0.6075} & \STYACC{0.4523} / \STYAUC{0.6206} & \STYACC{0.4982} / \STYAUC{0.6631} & \STYACC{0.4523} / \STYAUC{0.6585} & \STYACC{0.4452} / \STYAUC{0.6318} & \STYACC{0.4488} / \STYAUC{0.6372} & \STYACC{0.4134} / \STYAUC{0.5967} & \STYACC{0.4452} / \STYAUC{0.6145} \\
 & \texttt{CONTACT} & \STYACC{0.4099} / \STYAUC{0.5705} & \STYACC{0.4205} / \STYAUC{0.5428} & \STYACC{0.4452} / \STYAUC{0.5796} & \STYACC{0.4982} / \STYAUC{0.6485} & \STYACC{0.4558} / \STYAUC{0.6334} & \STYACC{0.4912} / \STYAUC{0.6289} & \STYACC{0.4735} / \STYAUC{0.6403} & \STYACC{0.4382} / \STYAUC{0.6330} & \STYACC{0.4735} / \STYAUC{0.6314} \\
 & \texttt{LateMIL} & \STYACC{0.4700} / \STYAUC{0.6027} & \STYACC{0.4912} / \STYAUC{0.6318} & \STYACC{0.4700} / \STYAUC{0.6169} & \STYACC{0.4629} / \STYAUC{0.6342} & \STYACC{0.4876} / \STYAUC{0.6346} & \STYACC{0.4841} / \STYAUC{0.6054} & \STYACC{0.4629} / \STYAUC{0.6301} & \STYACC{0.5159} / \STYAUC{0.6598} & \STYACC{0.4452} / \STYAUC{0.6315} \\
\hline
\end{tabular}%
}
\end{table*}

%% file: miscs/tables/full_conformal_table.tex
\begin{table}[t]
\centering
\caption{Conformal prediction metrics per task and model family across coverage levels ($\alpha \in \{0.05, 0.10, 0.20\}$). Each $\alpha$-block reports: empirical coverage (Cov.), coverage gap $\Delta = \text{cov}-(1-\alpha)$ (Gap), average prediction-set size (Set), and singleton rate (Sing.). Fixed columns (right): expected calibration error (ECE), multi-class Brier score, and top-1 accuracy (Acc.), which are independent of $\alpha$. Results averaged over encoder configurations.}
\label{tab:conformal_alpha_sweep}
\scriptsize
\setlength{\tabcolsep}{3pt}
\renewcommand{\arraystretch}{1.12}
\resizebox{\linewidth}{!}{%
\begin{tabular}{ll|cccc|cccc|cccc|ccc}
\hline
\multirow{2}{*}{Task} & \multirow{2}{*}{Model} & \multicolumn{4}{c}{$\alpha=0.05$} & \multicolumn{4}{c}{$\alpha=0.10$} & \multicolumn{4}{c}{$\alpha=0.20$} & ECE & Brier & Acc. \\
 & & Cov. & Gap & Set & Sing. & Cov. & Gap & Set & Sing. & Cov. & Gap & Set & Sing. & & & \\
\hline
\multirow{5}{*}{BC -- LOH} & CONTACT & 0.9536 & +0.0036 & 2.35 & 0.097 & 0.8955 & -0.0045 & 1.97 & 0.249 & 0.7898 & -0.0102 & 1.53 & 0.500 & 0.0575 & 0.510 & 0.609 \\
 & HEMIL & 0.9643 & +0.0143 & 2.48 & 0.081 & 0.9010 & +0.0010 & 2.02 & 0.251 & 0.8017 & +0.0017 & 1.55 & 0.499 & 0.0661 & 0.510 & 0.613 \\
 & LateMIL & 0.9603 & +0.0103 & 2.25 & 0.138 & 0.9081 & +0.0080 & 1.97 & 0.271 & 0.8071 & +0.0071 & 1.57 & 0.504 & 0.1444 & 0.545 & 0.599 \\
 & MCAT & 0.9515 & +0.0015 & 2.22 & 0.093 & 0.9014 & +0.0014 & 1.98 & 0.194 & 0.7878 & -0.0122 & 1.56 & 0.454 & 0.0620 & 0.515 & 0.579 \\
 & GeneMLP & 0.9578 & +0.0078 & 2.37 & 0.049 & 0.9060 & +0.0060 & 2.08 & 0.158 & 0.8029 & +0.0029 & 1.68 & 0.397 & 0.1596 & 0.593 & 0.573 \\
\hline
\multirow{5}{*}{BC -- Subtype} & CONTACT & 0.9481 & -0.0019 & 2.36 & 0.178 & 0.8981 & -0.0019 & 1.87 & 0.350 & 0.7947 & -0.0053 & 1.38 & 0.638 & 0.0531 & 0.463 & 0.654 \\
 & HEMIL & 0.9462 & -0.0038 & 2.73 & 0.068 & 0.9120 & +0.0120 & 2.34 & 0.163 & 0.8072 & +0.0072 & 1.70 & 0.424 & 0.1242 & 0.559 & 0.593 \\
 & LateMIL & 0.9484 & -0.0016 & 2.55 & 0.187 & 0.8969 & -0.0031 & 1.99 & 0.347 & 0.8100 & +0.0100 & 1.51 & 0.576 & 0.1250 & 0.507 & 0.646 \\
 & MCAT & 0.9553 & +0.0052 & 2.57 & 0.067 & 0.9059 & +0.0059 & 1.93 & 0.315 & 0.8027 & +0.0027 & 1.40 & 0.621 & 0.0549 & 0.466 & 0.652 \\
 & GeneMLP & 0.9562 & +0.0062 & 2.60 & 0.108 & 0.9025 & +0.0025 & 2.11 & 0.247 & 0.7992 & -0.0008 & 1.54 & 0.556 & 0.1240 & 0.523 & 0.621 \\
\hline
\multirow{5}{*}{NSCLC -- Biopsy Site} & CONTACT & 0.9642 & +0.0142 & 2.71 & 0.002 & 0.9159 & +0.0159 & 2.52 & 0.007 & 0.8256 & +0.0256 & 2.17 & 0.044 & 0.0508 & 0.630 & 0.453 \\
 & HEMIL & 0.9663 & +0.0163 & 2.72 & 0.003 & 0.9402 & +0.0402 & 2.55 & 0.015 & 0.8399 & +0.0399 & 2.10 & 0.132 & 0.1195 & 0.651 & 0.482 \\
 & LateMIL & 0.9776 & +0.0276 & 2.81 & 0.006 & 0.9420 & +0.0420 & 2.62 & 0.025 & 0.8574 & +0.0574 & 2.23 & 0.100 & 0.1482 & 0.671 & 0.455 \\
 & MCAT & 0.9686 & +0.0186 & 2.69 & 0.000 & 0.9325 & +0.0325 & 2.49 & 0.007 & 0.8279 & +0.0279 & 2.09 & 0.054 & 0.0436 & 0.623 & 0.463 \\
 & GeneMLP & 0.9591 & +0.0091 & 2.78 & 0.008 & 0.9156 & +0.0156 & 2.61 & 0.017 & 0.8193 & +0.0193 & 2.26 & 0.066 & 0.1349 & 0.695 & 0.406 \\
\hline
\multirow{5}{*}{NSCLC -- Tumor Site} & CONTACT & 0.9598 & +0.0098 & 2.72 & 0.005 & 0.9149 & +0.0149 & 2.51 & 0.031 & 0.8240 & +0.0240 & 2.16 & 0.105 & 0.0598 & 0.636 & 0.433 \\
 & HEMIL & 0.9693 & +0.0193 & 2.75 & 0.010 & 0.9258 & +0.0258 & 2.48 & 0.047 & 0.8174 & +0.0174 & 2.03 & 0.147 & 0.0754 & 0.623 & 0.474 \\
 & LateMIL & 0.9593 & +0.0093 & 2.77 & 0.014 & 0.9098 & +0.0098 & 2.51 & 0.046 & 0.8073 & +0.0073 & 2.10 & 0.156 & 0.1231 & 0.674 & 0.444 \\
 & MCAT & 0.9614 & +0.0114 & 2.68 & 0.004 & 0.9216 & +0.0216 & 2.49 & 0.015 & 0.8298 & +0.0298 & 2.13 & 0.079 & 0.0679 & 0.633 & 0.453 \\
 & GeneMLP & 0.9471 & -0.0029 & 2.77 & 0.009 & 0.8986 & -0.0014 & 2.58 & 0.023 & 0.8150 & +0.0150 & 2.26 & 0.082 & 0.2037 & 0.773 & 0.397 \\
\hline
\end{tabular}%
}
\end{table}